\documentclass[a4paper,11pt]{scrartcl}

\usepackage{amsmath,amssymb,amsthm}
\usepackage{authblk}
\usepackage{color}
\usepackage[top=2.3cm,bottom=2.3cm,left=2.5cm,right=2.5cm]{geometry}
\usepackage{graphicx}
\usepackage{hyperref}
\usepackage[utf8]{inputenc}
\usepackage{pdfpages}
\usepackage{url}

\newcommand{\dd}{\operatorname{d}\!}
\DeclareMathOperator\erf{erf}

\title{Generalisation dynamics of online learning in over-parameterised neural networks}

\author[1]{Sebastian Goldt}
\author[2]{Madhu S.~Advani}
\author[3]{Andrew M.~Saxe}
\author[4]{\\Florent Krzakala}
\author[1]{Lenka Zdeborov\'a}
\affil[1]{Institut de Physique Th\'eorique, CNRS, CEA, Universit\'e
  Paris-Saclay, France}
\affil[2]{Center for Brain Science, Harvard University,
  Cambridge, MA 02138, USA}
\affil[3]{Department of Experimental Psychology, University of
  Oxford, United Kingdom}
\affil[4]{Laboratoire de Physique de l’Ecole Normale Sup\'erieure, Universit\'e
  PSL, CNRS, Sorbonne Universit\'e, Universit\'e Paris-Diderot, Sorbonne Paris
  Cit\'e, Paris, France}
 \date{}

\begin{document}
\maketitle

\begin{abstract}
  Deep neural networks achieve stellar generalisation on a variety of problems,
  despite often being large enough to easily fit all their training data.
  Here we study the generalisation dynamics of two-layer neural networks in a
  teacher-student setup, where one network, the student, is trained using
  stochastic gradient descent (SGD) on data generated by another network, called
  the teacher.
  We show how for this problem, the dynamics of SGD are captured by a set of
  differential equations.
  In particular, we demonstrate analytically that the generalisation error of the
  student increases linearly with the network size, with other relevant parameters 
  held constant.
  Our results indicate that achieving good generalisation in neural networks depends on 
  the interplay of at least the algorithm, its learning rate, 
  the model architecture, and the data set.
\end{abstract}

\section{Introduction}\label{sec:intro}

One hallmark of the deep neural networks behind state-of-the-art results in
image classification~\cite{Lecun2015} or the games of Atari and
Go~\cite{Mnih2015,Silver2016} is their size: their free parameters outnumber the
samples in their training set by up to two orders of
magnitude~\cite{Simonyan2015}. Statistical learning theory would suggest that
such heavily over-parameterised networks should generalise poorly without
further
regularisation~\cite{Bartlett2003,Mohri2012,Neyshabur2015a,Golowich2017,Dziugaite2017,Arora2018,Allen-Zhu2018},
yet empirical studies consistently find that increasing the size of networks to
the point where they can fit their training data and beyond does not impede
their ability to generalise well~\cite{Neyshabur2015,Zhang2016a,Arpit2017}. This
paradox is arguably one of the biggest challenges in the theory of deep
learning.


In practice, it is notoriously difficult to determine the point at which a
statistical model becomes over-parameterised for a given data set. Instead, here
we study the dynamics of learning neural networks in the teacher-student
setup. The student is a two-layer neural network with weights $w$ that computes
a scalar function $\phi(w, x)$ of its inputs $x$. It is trained with samples
$(x, y)$, where $y=\phi(B, x) + \zeta$ is the noisy output of another two-layer
network with weights $B$, called the teacher, and $\zeta$ is a Gaussian random
variable with mean 0 and variance~$\sigma^2$. Crucially, the student can have a
number of hidden units $K$ that is different from $M$, the number of hidden
units of the teacher. Choosing $L \equiv K-M>~0$ then gives us neural networks
that are over-parameterised with respect to the generative model of their
training data in a controlled way. The key quantity in our model is the
generalisation error
\begin{equation}
  \label{eq:eg}
  \epsilon_g \equiv \frac{1}{2} \langle {\left[\phi(w, x)-\phi(B,
    x)\right]}^2\rangle,
\end{equation}
where the average $\langle \cdot \rangle$ is taken over the input
distribution. Our main questions are twofold: how does $\epsilon_g$ evolve over
time, and how does it depend on~$L$?

\textbf{Main contributions.} We derive a set of ordinary differential
equations (ODEs) that track the \emph{typical} generalisation error of an 
over-parameterised student
trained using SGD. This description becomes exact for large input dimension and
data sets that are large enough to allow that we visit every sample only once
before training converges. Using this framework, we analytically calculate the
generalisation error after convergence $\epsilon_g^*$. We find that
with other relevant parameters held constant, the generalisation error 
increases at least
linearly with~$L$. For small learning rates $\eta$ in particular, we have
\begin{equation}
  \label{eq:eg_scaling}
  \epsilon_g^* \sim \eta \sigma^2 L.
\end{equation}
Our model thus offers an interesting perspective on the implicit regularisation
of SGD, which we will discuss in detail. The derivation of a set of ODEs for
over-parameterised neural networks and their perturbative solution in the limit
of small noise are an extension of earlier work by~\cite{Biehl1995}
and~\cite{Saad1995a,Saad1995b}.

The concepts and tools from statistical physics that we use in our analysis have
a long and successful history of analysing average-case generalisation in
learning and
inference~\cite{Gardner1989,Seung1992,Watkin1993,Engel2001,Zdeborova2016}, and
they have recently seen a surge of
interest~\cite{Advani2016,Chaudhari2017,Advani2017,Aubin2018,Baity-Jesi2018}.

We begin our paper in Sec.~\ref{sec:setup} with a description of the
teacher-student setup, the learning algorithm and the derivation of a set of
ordinary differential equations that capture the dynamics of SGD in our
model. Using this framework, we derive Eq.~\eqref{eq:eg_scaling} in
Sec.~\ref{sec:final-eg} and discuss networks with sigmoidal, linear and ReLU
activation functions in detail. We discuss our results and in particular the
importance of the size of the training set in Sec.~\ref{sec:discussion} before
concluding in Sec.~\ref{sec:conclusion}.

\section{Setup}\label{sec:setup}

\subsection{The teacher generates test and training data}\label{sec:data}

\begin{figure}
  \centering
  \includegraphics[width=.5\linewidth]{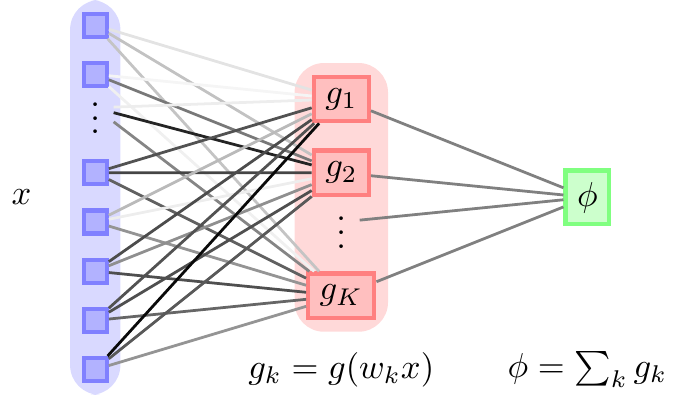}
  \caption{\label{fig:architecture} \textbf{Neural network with a single hidden
      layer.} A network with $K$ hidden units and weights $w$ implements a
    scalar function of its inputs $x$, $y=\sum_k^K g(w_k x)$, where
    $g: \mathbb{R}\to\mathbb{R}$ is the non-linear activation function of the
    network.}
\end{figure}

We study the learning of a supervised regression problem with inputs
$x\in\mathbb{R}^N$ and outputs $y(x)\in\mathbb{R}$ with a generative model as
follows. We take the components of the inputs $x_n$ ($n=1,\ldots,N$) to be
i.i.d.\ Gaussian random variables\footnote{N.B. our results for large $N$ are
  valid for any input distributions that has the same mean and variance, for
  example equiprobable binary inputs $x_n=\pm1$.} with zero mean and unit
variance. The output $y(x)$ is given by a neural network with a single hidden
layer containing $M$ hidden units and all-to-all connections, see
Fig.~\ref{fig:architecture}. Its weights $B\in\mathbb{R}^{M\times N}$ from the
inputs to the hidden units are drawn at random from some distribution $p(B)$ and
kept fixed\footnote{It is also possible to extend our approach to
  time-dependent weights $B$.}. Given an input $x$, the network's output is given by
\begin{equation}
  \label{eq:phi}
  \phi(B, x) = \sum_{m=1}^M g\left(\frac{B_m x}{\sqrt{N}} \right)
\end{equation}
where $B_m$ is the $m$th row of $B$, $B_m x$ is the dot product between two
vectors and $g: \mathbb{R} \to \mathbb{R}$ is the activation function of the
network. We focus on the case where both student and teacher have the same
activation function. In particular, we study linear networks with $g(x)=x$,
sigmoidal networks with $g(x) = \erf\left(x/\sqrt{2}\right)$, and rectified
linear units where $g(x) = {\max}(0, x)$. The training set consists of $P$
tuples $(x^\mu, y^\mu_B(x^\mu))$, $\mu=1,\ldots,P$; where
\begin{equation}
  \label{eq:y}
  y^\mu_B(x^\mu) \equiv \phi(B, x^\mu) + \zeta^\mu,
\end{equation}
is a noisy observation of the network's output and the random variable
$\zeta^\mu$ is normally distributed with mean $0$ and variance $\sigma^2$.

In the statistical physics literature on learning, a data-generating neural
network is called the \emph{teacher} and neural networks of the
type~\eqref{eq:phi} are called Soft Committee
Machines~\cite{Schwarze1993a}. They combine the apparent simplicity of their
architecture, which allows for a detailed analytical description, with the power
of a universal approximator: given a hidden layer of sufficient size, they can
approximate any continuous function of their inputs to any desired
accuracy~\cite{Cybenko1989,Hornik1989}. They have thus been at the center of a
lot of recent research on the generalisation of neural
networks~\cite{Mei2018,Rotskoff2018,Aubin2018,Chizat2018,Li2018a}.

\subsection{The student aims to mimic the teacher's function}

Once a teacher network has been chosen at random, we train another neural
network, called the \emph{student}, on data generated by the teacher according
to~\eqref{eq:y}. The student has a single fully-connected hidden layer with $K$
hidden units, and we explicitly allow for $K\neq M$. The student's weights from
the input to the hidden layer are denoted $w\in\mathbb{R}^{K \times N}$ and its
output is given by $\phi(w, x)$. We keep the weights from the hidden units to
the output fixed at unity and only train the first layer weights $w$. 
We consider both networks in the thermodynamic limit, where we
let $N\to\infty$ while keeping $M, K$ of order~1.

The key quantity in our study is the generalisation error of the student network
with respect to the teacher network, which we defined in Eq.~\eqref{eq:eg} as
the mean squared error between the outputs of the student and the noiseless
output of the teacher, averaged over the distribution of inputs. Note that
including the output noise of the teacher would only introduce a constant offset
proportional to the variance of the output noise.

\subsection{The student is trained using online learning}

Since we are training the student for a regression task, we choose a quadratic
loss function. Given a training data set with $P$ samples, the training error
reads
\begin{equation}
  \label{eq:et}
  E(w) =\frac{1}{2} \sum_{\mu=1}^P {\left( \phi(w, x^\mu) - y_B^\mu\right)}^2.
\end{equation}
We perform stochastic gradient descent on the training error to optimise the
weights of the student, using only a single sample $(x^\mu, y_B^\mu)$ to
evaluate the gradient of $E(w)$ at every step. To make the problem analytically
tractable, we consider the limit where the training data set is large
enough to allow that each sample $(x^\mu$, $y_B^\mu)$ is visited only once
during the entire training until the generalisation error converges to its final
value\footnote{In Sec.~\ref{sec:discussion}, we investigate the case of
  small~$P$ via simulations.}. We can hence index the steps of the algorithm by
$\mu$ and write the weight updates as
\begin{equation}
  w_k^{\mu+1} = w_k^{\mu} - \frac{\kappa}{N} w_k^\mu - \frac{\eta}{\sqrt{N}}
                 \nabla_{w_k} E(w) |_{(x^\mu, y_B^\mu)}
\end{equation}
where $\kappa$ is the weight decay rate, the learning rate is $\eta$, and we
have chosen their scaling with $N$ such that all terms remain of order~1 in the
thermodynamic limit $N\to\infty$. Evaluating the derivative yields
\begin{equation}
  \label{eq:sgd}
  w_k^{\mu+1} = w_k^{\mu} - \frac{\kappa}{N} w_k^\mu - \frac{\eta}{\sqrt{N}}
  x^\mu r_k^\mu
\end{equation}
where
\begin{equation}
  \label{eq:delta}
  r_k^\mu \equiv g'(\lambda_k^\mu) \left[ \phi(w, x^\mu)- y_B^\mu\right]
\end{equation}
and we have defined $\lambda_k^\mu \equiv w_k x^\mu / \sqrt{N}$. 

Stochastic gradient descent with mini-batch size 1 in the limit of very large
training sets is also known as online or one-shot
learning. Kinzel~\cite{Kinzel1990} first realised that its dynamics could be
described in terms of order parameters, initiating a number of works on the
perceptron~\cite{Mace1998,Engel2001}. Online learning in committee machines was
first studied by Biehl and Schwarze~\cite{Biehl1995} in the case $M=1, K=2$ and
by Saad and Solla~\cite{Saad1995b,Saad1997b}, who gave a detailed analytic
description of the case $K=M$ with $g(x)=\erf\left(x/\sqrt{2}\right)$. Beyond
its application to neural networks, its performance has been analysed for
problems ranging from PCA~\cite{Oja1985,Wang2017} to the training of generative
adversarial networks~\cite{Wang2018}.

\subsection{The dynamics of online learning can be described in closed form}

\begin{figure}[t!]
  \centering
  \includegraphics[width=.33\linewidth]{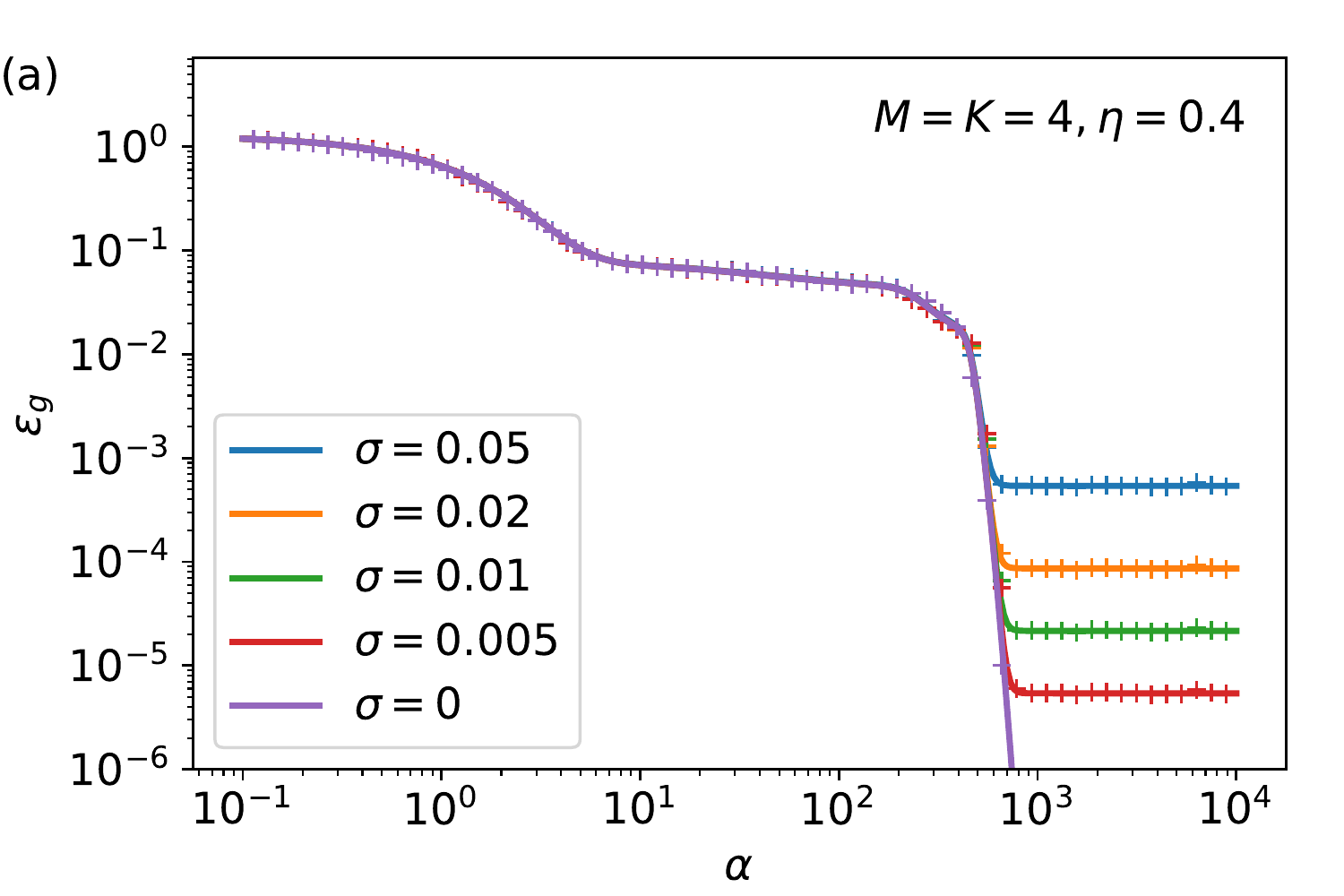}%
  \includegraphics[width=.33\linewidth]{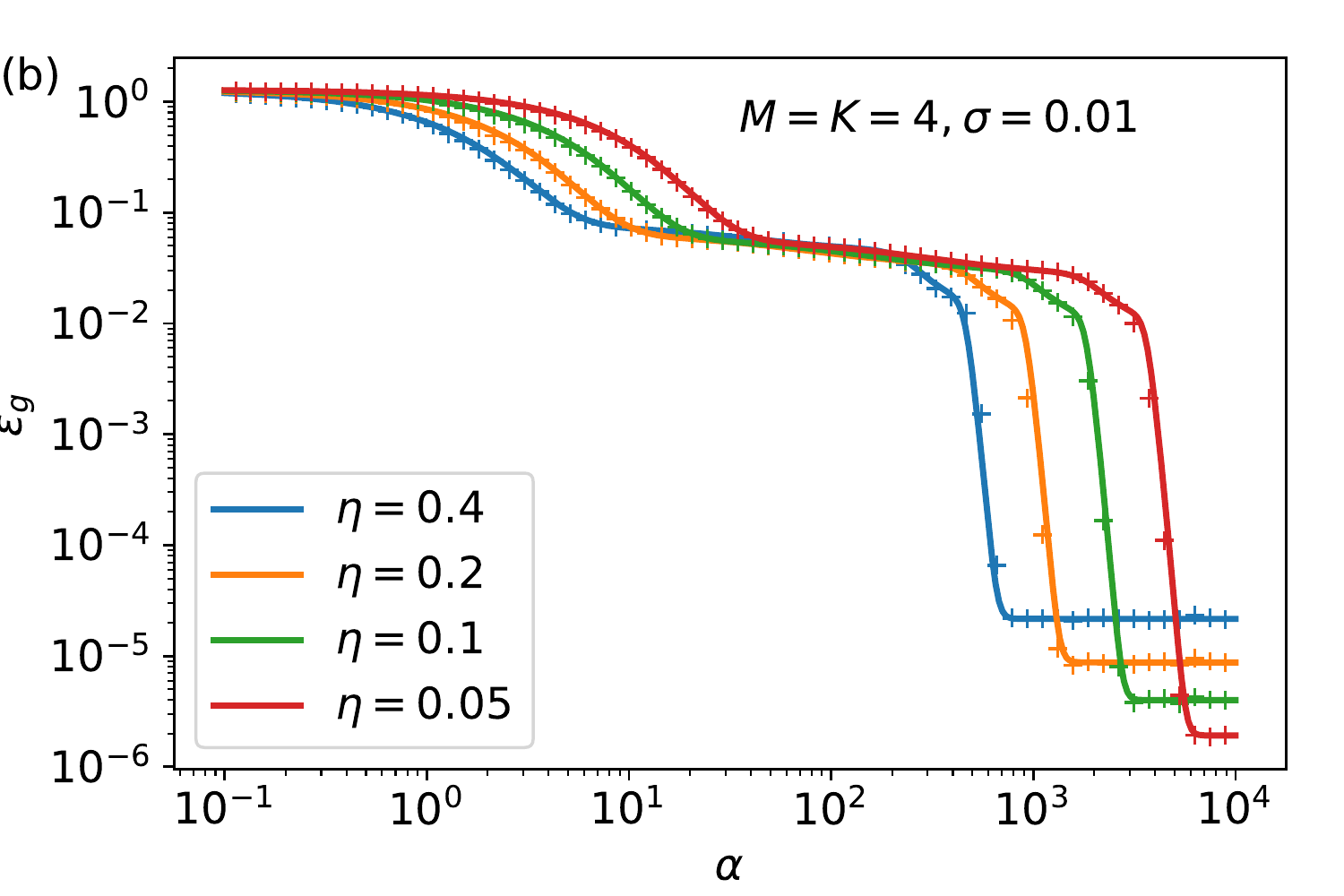}%
  \includegraphics[width=.33\linewidth]{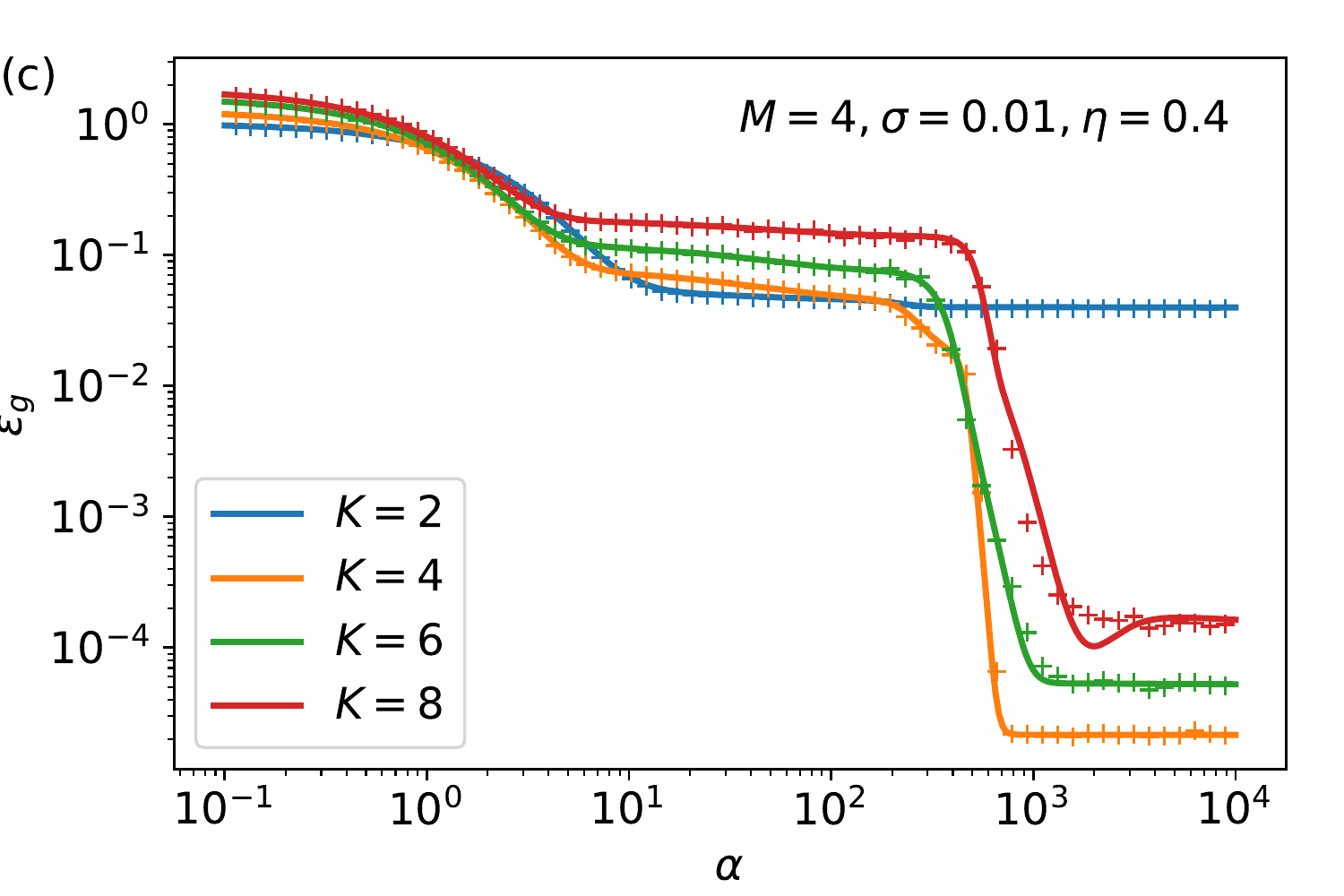}
  \caption{\label{fig:eg_erf_scaling} \textbf{The analytical description of the
      generalisation dynamics of sigmoidal networks (solid) matches simulations
      (crosses).} We show learning curves $\epsilon_g(\alpha)$ obtained by
    integration of the ODEs~\eqref{eq:eom} (solid). From left to the right, we
    vary the variance of the teacher's output noise $\sigma$, the learning rate
    $\eta$, and the number of hidden units in the student $K$. For each
    combination of parameters shown in the plots, we ran a single simulation of
    a network with $N=784$ and plot the generalisation observed (crosses). $\kappa=0$ in all cases.}
\end{figure}

Our aim is to track the evolution of the generalisation
error~$\epsilon_g$~\eqref{eq:eg}, which can be written more explicitly as
\begin{equation}
  \label{eq:eg-explicit}
  \epsilon_g = \frac{1}{2}\left\langle {\left[ \sum_{k=1}^K
        g\left(\lambda_k^\mu\right) - \sum_{m=1}^M g(\nu_m^\mu)\right]}^2 \right\rangle,
\end{equation}
where $\nu_m^\mu \equiv B_m x^\mu/\sqrt{N}$. Since the inputs $x$ only appear as
products with the weight vectors of the student and the teacher, we can replace
the average~$\langle \cdot \rangle$ over~$x$ with an average
over~$\lambda_k^\mu$ and~$\nu_m^\mu$. To determine the distribution of the
latter, the assumption that every sample $(x^\mu, y_B^\mu)$ is only used once
during training becomes crucial, because it guarantees that the inputs and the
weights of the networks are uncorrelated. By the central limit theorem,
$\nu_m^\mu, \lambda_k^\mu$ are hence normally distributed with mean zero since
$\langle x_n \rangle=0$. Their covariance is also easily found: writing $w_{ka}$
for a component of the $k$th weight vector, we have
\begin{equation}
  \label{eq:Q}
  \langle \lambda_k \lambda_l \rangle = \frac{\sum_{a,b}^N w_{ka} w_{lb}\langle
    x_a x_b \rangle}{N}=\frac{w_k w_l}{N} \equiv Q_{kl}
\end{equation}
since $\langle x_a x_b \rangle=\delta_{ab}$. Likewise, we define
\begin{equation}
  \label{eq:RandT}
  \langle \nu_n \nu_m \rangle = \frac{B_n B_m}{N} \equiv T_{nm}, \quad
  \langle \lambda_k \nu_m \rangle = \frac{w_k B_m}{N} \equiv R_{km}.
\end{equation}
The variables $R_{in}$, $Q_{ik}$, and $T_{nm}$ are called \emph{order
  parameters} in statistical physics and measure the overlap between student and
teacher weight vectors $w_i$ and $B_n$ and their self-overlaps,
respectively. Crucially, from Eq.~\eqref{eq:eg-explicit} we see that they are
sufficient to determine the generalisation error $\epsilon_g$.

We can obtain a closed set of differential equations for the time evolution of
the order parameters $Q$ and $R$ by squaring the weight update~\eqref{eq:sgd}
and taking its inner product with $B_n$, respectively\footnote{Since we keep the
  teacher fixed, $T$ remains constant; however, our approach can be easily
  extended to a time-dependent teacher.}. Then, an average over the inputs $x$
needs to be taken. The resulting equations of motion for $R$ and $Q$ can be
written as
\begin{subequations}\label{eq:eom}
  \begin{align}
    \frac{\dd R_{in}}{\dd \alpha} &= -\kappa R_{in} + \eta \langle r_i \nu_n \rangle\\
    \begin{split}
      \frac{\dd Q_{ik}}{\dd \alpha} &= -2 \kappa Q_{ik} + \eta \langle r_i
      \lambda_k \rangle + \eta \langle r_k \lambda_i \rangle\\
      & \qquad + \eta^2 \langle r_i r_k\rangle + \eta^2 \sigma^2 \langle
      g'(\lambda_i)g'(\lambda_k) \rangle
    \end{split}
  \end{align}
\end{subequations}
where $\alpha=\mu / N$ becomes a continuous time-like variable in the limit
$N\to\infty$. The averages over inputs $\langle \cdot \rangle$ can again be
reduced to an average over the normally distributed local fields $\lambda_k^\mu$
and $\nu_m^\mu$ as above. If the averages can be evaluated analytically, the
equations of motion~\eqref{eq:eom} together with the generalisation
error~\eqref{eq:eg-explicit} form a closed set of equations which can be
integrated numerically and provides an exact description of the generalisation
dynamics of the network in the limit of large $N$ and large training
sets. Indeed, the integrals have an analytical solution for the choice
$g(x)=\erf(x/\sqrt{2})$~\cite{Biehl1995} and for linear
networks. Eqns.~\eqref{eq:eom} hold for any $M$ and $K$, enabling us to study
the learning of complex non-linear target functions $\phi(B, x)$, rather than
data that is linearly separable or follows a Gaussian
distribution~\cite{Brutzkus2018,Soltanolkotabi2018}. A detailed derivation and
the explicit form of the equations of motion are given in
Appendix~\ref{sec:app-dynamics}.

We plot $\epsilon_g(\alpha)$ obtained by numerically integrating\footnote{We
  have packaged our simulations and our ODE integrator into a user-friendly
  Python library. To download, visit \url{https://github.com/sgoldt/pyscm}}
Eqns.~\eqref{eq:eom} and the generalisation error observed during a single run
of online learning~\eqref{eq:sgd} with $N=784$ in
Fig.~\ref{fig:eg_erf_scaling}. The plots demonstrate a good quantitative
agreement between simulations and theory and display some generic features of
online learning in soft committee machines.

One notable feature of all the plots in Fig.~\ref{fig:eg_erf_scaling} is the
existence of plateaus during training, where the generalisation error is almost
stationary. During this time, the student ``believes'' that data are linearly
separable and all its hidden units have roughly the same overlap with all the
hidden units of the teacher. Only after a longer time, the student picks up the
additional structure of the teacher and ``specialises'': each of its hidden
units ideally becomes strongly correlated with one and only one hidden unit of
the teacher before the generalisation error decreases exponentially to its final
value.  This effect is well-known in the literature for both batch and online
learning~\cite{Schwarze1993a,Biehl1995,Saad1995a} and will be revisited in
Sec.~\ref{sec:final-eg}.

It is perhaps surprising that the generalisation dynamics seem unaffected by the
difference in output noise (Fig.~\ref{fig:eg_erf_scaling} a) until they leave
the plateau. This is due to the fact that the noise appears in the equations of
motion only in terms that are quadratic in the learning rate. Their effect takes
longer to build up and become significant. The specialisation observed above is
also due to terms that are quadratic in the learning rate. This goes to show
that even in the limit of small learning rates, one cannot simplify the dynamics
of the neural network by linearising Eqns.~\eqref{eq:eom} in $\eta$ without
losing some key properties of the dynamics.

\section{Asymptotic generalisation of over-parameterised students after online
  learning}\label{sec:final-eg}

In the absence of output noise ($\sigma=0$) and without weight decay
($\kappa=0$), online learning of a student with $K\ge M$ hidden units will yield
a network that generalises perfectly with respect to the teacher. More
precisely, at some point during training, the generalisation error will start an
exponential decay towards zero (see Appendix~\ref{sec:app-no-noise}). On the
level of the order parameters $Q$ and $R$, a student that generalises perfectly
with respect to the teacher corresponds to a stable fixed point of the equations
of motion~\eqref{eq:eom} with $\epsilon_g=0$.

This fixed point disappears for $\sigma>0$ and the order parameters converge to
a different fixed point. The values of the order parameters at that fixed point
can be obtained perturbatively in the limit of small noise, \emph{i.e}.\  small
$\sigma$. To this end, we first make an ansatz for the matrices $Q$ and $R$ that
involves eight order parameters for any $M, K$. We choose this number of order 
parameters for two reasons: it is the smallest number of
variables for which we were able to self-consistently close the equations of
motion~\eqref{eq:eom}, and they agree with numerical evidence obtained from
integrating the full equations of motion~\eqref{eq:eom}.

We then derive equations of 
motion for this reduced set of order parameters, and expand them to first order
in $\sigma$ around the fixed point with perfect generalisation. Throughout this
section, we set $\kappa=0$ and choose uncorrelated and isotropic weight vectors
for the teacher, \emph{i.e.}  $T_{nm}=\delta_{nm}$, which is equivalent to
drawing the weights at random from a standard normal distribution.

\subsection{Sigmoidal networks}\label{sec:sigmoidal-network}

\begin{figure}[t!]
  \centering
  \includegraphics[width=.5\linewidth]{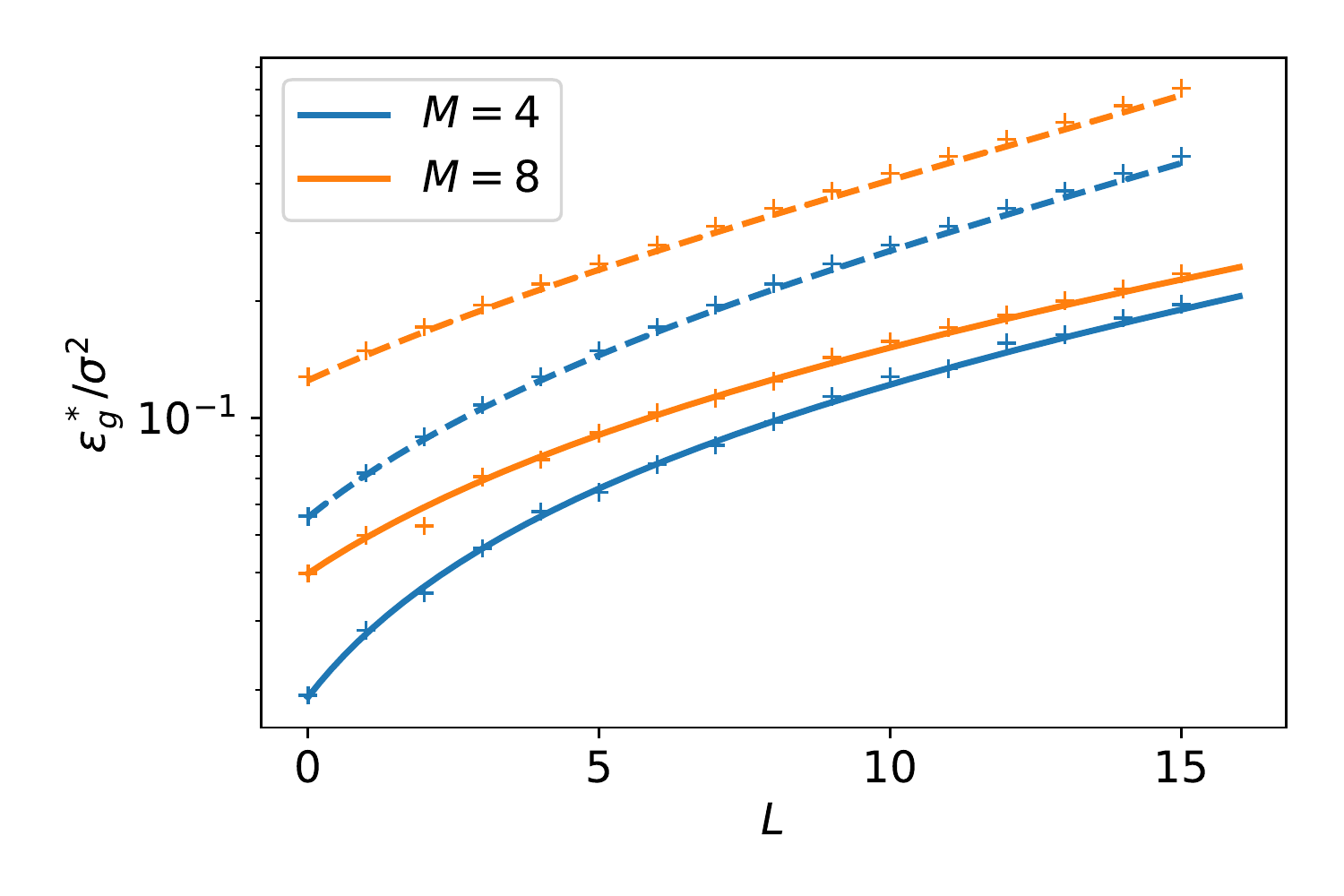}
  \caption{\label{fig:eg_vs_L} \textbf{Theoretical predictions for
      $\epsilon_g^*$ match simulations.} We plot theoretical predictions for
    $\epsilon_g^*/\sigma^2$ for sigmoidal networks (Eq.~\eqref{eq:egFinal},
    solid line) and linear networks (Eq.~\eqref{eq:eg-lin}, dashed) together
    with the result from a single simulation of a network with
    $N=784$. Parameters: $\eta=0.05, \sigma=0.01$.}
\end{figure}

We have performed this calculation for teacher and student networks with
$g(x)=\erf(x/\sqrt{2})$. We discuss the details of this tedious calculation in
Appendix~\ref{sec:final-eg}, and here we only state the asymptotic value of the
generalisation error $\epsilon_g^*$ to first order in the variance of the noise
$\sigma^2$ for teacher and student with sigmoidal activation:
\begin{equation}
  \label{eq:egFinal}
  \epsilon_g^* = \frac{\sigma^2 \eta}{2 \pi} f(M, L, \eta) + \mathcal{O}(\sigma^3)
\end{equation}
where $f(M, L, \eta)$ is a lengthy rational function of its variables. The full
expression spans more than two pages, so here we plot it in
Fig.~\ref{fig:eg_vs_L} together with a single run of a simulation of a neural
network with $N=784$, which is in excellent agreement.

\subsubsection{Discussion}

One notable feature of Fig.~\ref{fig:eg_vs_L} is that with all else being equal,
the generalisation error increases monotonically with $L$. In other words, our
result~\eqref{eq:egFinal} implies that SGD alone fails to regularise the student
networks of increasing size in our setup, instead yielding students whose
generalisation error increases at least linearly with~$L$.

One might be tempted to mitigate this effect by simultaneously decreasing the
learning rate $\eta$ for larger students. However, this raises two problems:
first of all, a lower learning rate means the model would take longer to
train. More importantly though, the resulting longer training time implies that
more data is required until the final generalisation error is achieved. This is
in agreement with statistical learning theory, where given more and more data,
models with more parameters (higher $L$) and a higher complexity class,
\emph{e.g.} a higher VC dimension or Rademacher complexity~\cite{Mohri2012},
generalise just as well as smaller ones. In practice however, more data might
not be readily available. Furthermore, we show in the
Appendix~\ref{sec:app-eg-discussion} that even when we choose $\eta=1/K$, the
generalisation error still increases with $L$ before plateauing at a constant
value.

\begin{figure}[t!]
  \centering
  \includegraphics[width=.5\textwidth]{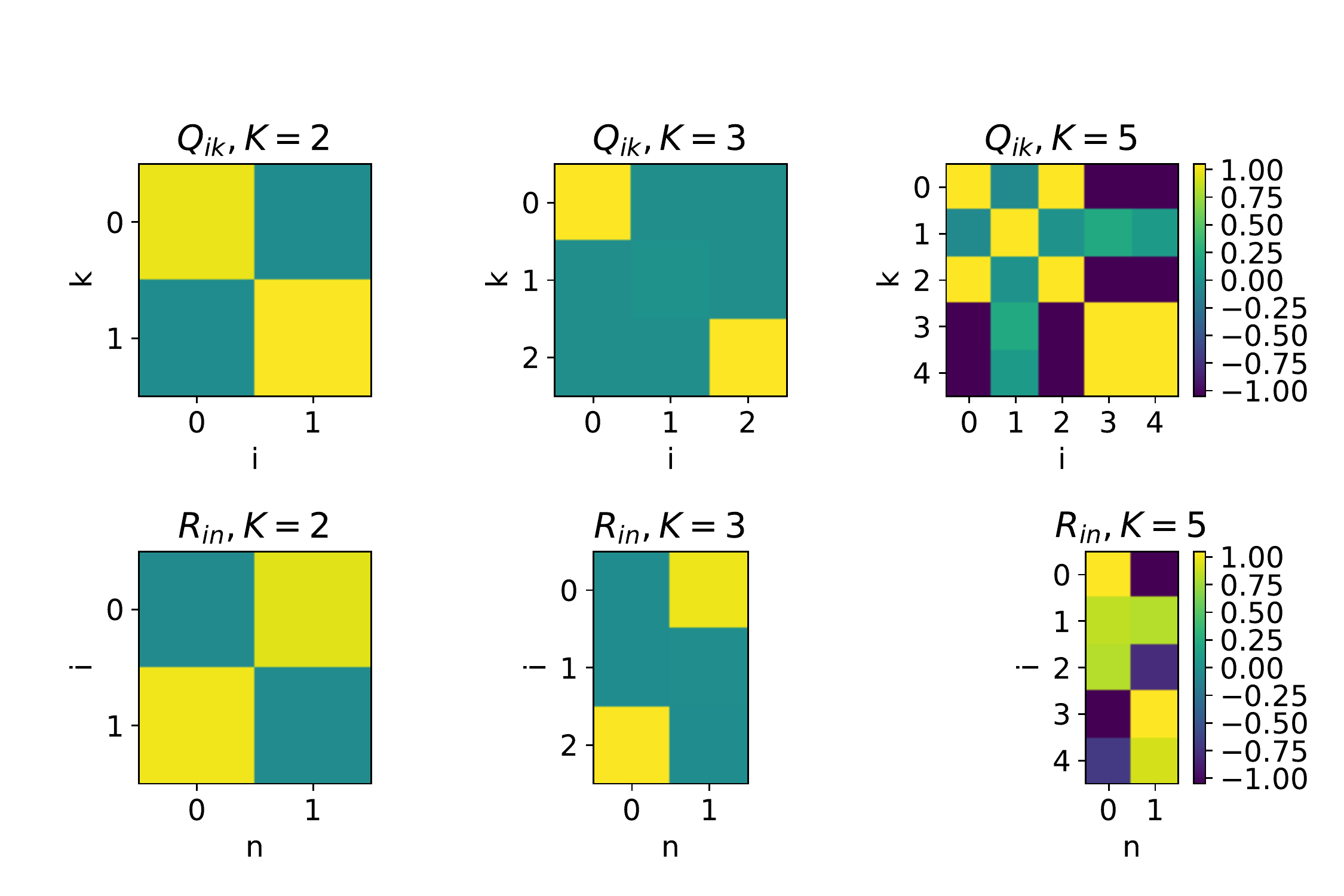}%
  \includegraphics[width=.5\textwidth]{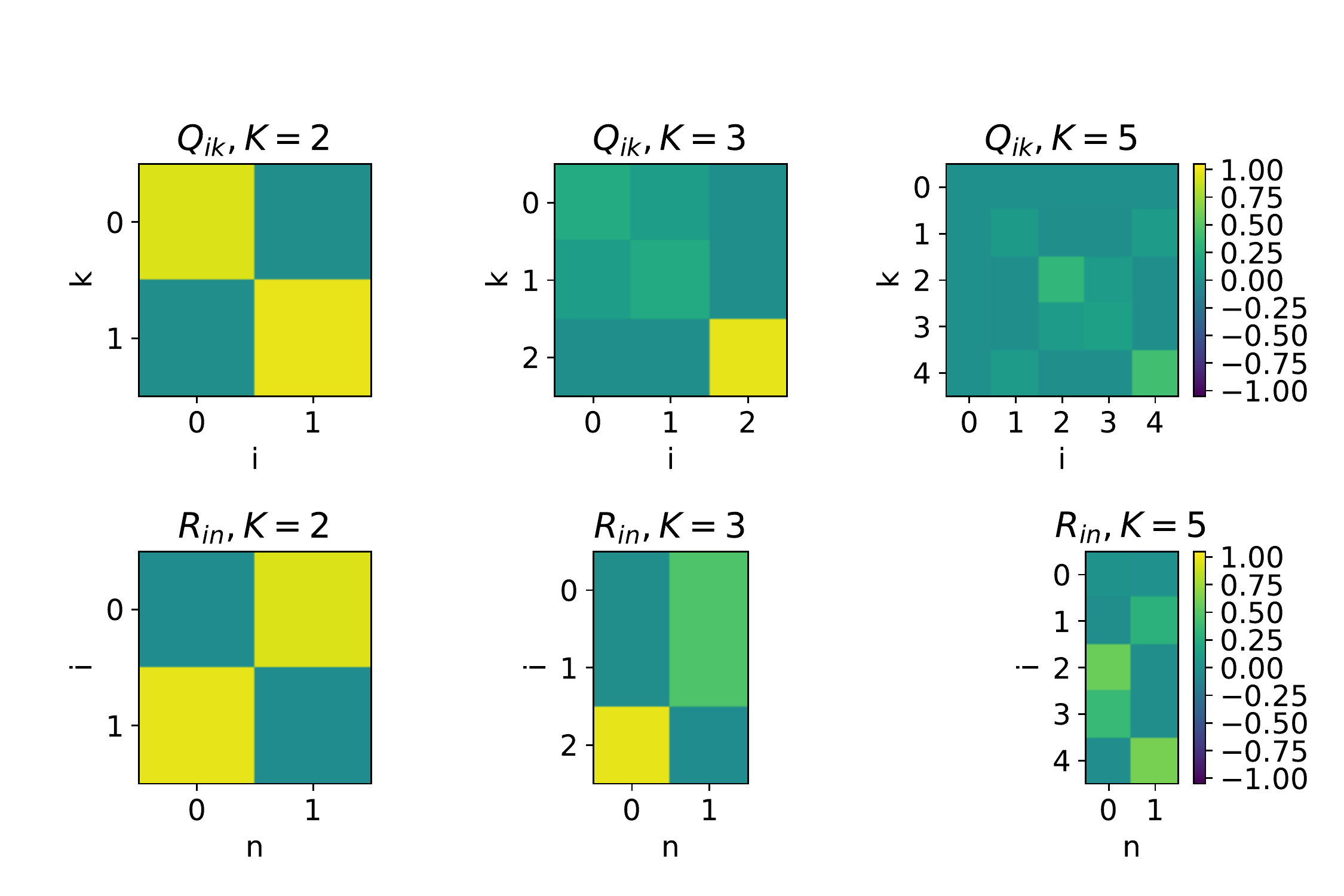}
  \caption{\label{fig:final-representations-noisy} \textbf{Sigmoidal networks
      learn different representations from a noisy teacher than ReLU networks.}
    For sigmoidal networks (\emph{left}), we see clear signs of specialisation
    as described in Sec.~\ref{sec:sigmoidal-network}: for $K=3$, one unit is
    simply shut down: $w_1=0$. As we increase $K$, some units become exactly
    anti-correlated to each other (e.g.\ units 1, 4 for $K=5$), hence
    effectively setting their weights to zero (we keep the weights of the second
    layer fixed at unity). ReLU networks instead find representations where
    several nodes are used. Parameters:
    $N=784, \eta=0.3, \sigma=0.1, \kappa=0$.}
\end{figure}

We can gain some intuition for the result~\eqref{eq:egFinal} by considering the
final representations learnt by a sigmoidal network. On the left half of
Fig.~\ref{fig:final-representations-noisy}, we plot the overlap matrices $Q$ and
$R$ for a teacher with $M=2$ and various~$K$. For $K=2$, we see that each of the
hidden units of the student has learnt the weights of one teacher hidden unit,
yielding diagonal matrices $Q$ and $R$ (modulo the permutation symmetry of the
hidden units). As we add a third hidden unit to the student, $K=3$, the
specialisation discussed in the previous section becomes apparent: two of the
hidden units of the student each align almost perfectly with a different hidden
unit of the teacher, such that $R_{01}=R_{20}\approx1$, while the weights of the
third unit go to zero ($Q_{11}\approx0$). As we add even more hidden units,
($K=5$), the weight vectors of some units become exactly anti-correlated, hence
effectively setting their weights to zero as far as the output of that network
is concerned (since we set the weights of the second layer to unity). This behaviour
is essentially a consequence of the sigmoidal form of the activation function, 
which makes it hard to express the sum of $M$ hidden units with $K\neq M$ hidden
units, instead forcing 1-to-1 specialisation of the student's hidden units.

In the over-parameterised case, $L=K-M$ units of the student are hence
effectively specialising to hidden units of the teacher with zero
weights. Although their weights are unbiased estimators of the weights $B=0$,
their variance is finite due to the noise in the teacher's output. Thus they
always hurt generalisation.

This intuition can be confirmed analytically by going to the limit of small
learning rates, which is the most relevant in practice. Expanding $\epsilon_g^*$
to first order in the learning rate reveals a particularly revealing form,
  \begin{equation}
    \label{eq:egFinal1stOrderInLr} \epsilon_g^* = \frac{\sigma^2 \eta}{2 \pi}
    \left(L + \frac{M}{\sqrt{3}} \right) + \mathcal{O}(\eta^2),
\end{equation}
with second-order corrections that are quadratic in $L$. The linear term in
$\eta$ is the sum of two contributions: the asymptotic generalisation errors of
$M$ independent networks with one hidden units that are learning from a teacher
with single hidden unit hand $T=1$. The $L=K-M$ superfluous units contribute
each the error of a continuous perceptrons that is learning from a teacher with
zero weights ($T=0$). Again, we relegate the detailed calculation to the
Appendix~\ref{sec:app-cp}.

\subsection{Linear networks}\label{sec:linear-networks}

One might suspect that part of the scaling $\epsilon_g^*\sim L$ in sigmoidal
networks is due to the specialisation of the hidden units or the fact that
teacher and student network can implement functions of different range if
$K \neq M$. It thus makes sense to calculate $\epsilon_g^*$ for linear neural
networks, where $g(x)=x$~\cite{Krogh1992a}. These networks have no
specialisation transition~\cite{Aubin2018} and their output range is set by the
magnitude of their weights, rather than their number of hidden
units. Furthermore, linear networks are receiving increasing attention as models
for neural networks~\cite{Saxe2014,Advani2017,Lampinen2018}.

Following the same steps as for the sigmoidal networks, a perturbative
calculation in the limit of small noise yields
\begin{equation}
  \label{eq:eg-lin}
  \epsilon_g^* = \frac{\eta  \sigma ^2 (L+M)}{4 - 2\eta(L+M)} + \mathcal{O}(\sigma^3).
\end{equation}
In the limit of small learning rates, the above expression further simplifies to
\begin{equation}
  \label{eq:eg-lin-smallLr}
  \epsilon_g^* = \frac{1}{4} \eta  \sigma ^2 (L+M)+\mathcal{O}\left(\eta
    ^2\right).
\end{equation}
Hence we see that in the limit of small learning rates, the asymptotic
generalisation error of linear networks has the same scaling with $\sigma$,
$\eta$ and $L$ as for sigmoidal networks. This result is again in good agreement
with the results of simulations, demonstrated in Fig.~\ref{fig:eg_vs_L}.

\subsubsection{Discussion}

The linear scaling of $\epsilon_g^*$ with $L$, keeping all other things equal,
might be surprising given that all linear networks implement a linear
transformation with an effective matrix $W=\sum_k w_k$, irrespective of their
number of hidden units. However, this is exactly the problem: adding hidden
units to a linear network does not augment the class of functions it can
implement, but it adds free parameters which will indiscriminately pick up
fluctuations due to the output noise in the teacher. 
The optimal generalisation error is indeed realised with $K=1$ irrespective of $M$,
since a linear network with $K=1$ has the lowest number of free parameters while
having the same expressive power as a teacher with arbitrary $M$. Our
formula~\eqref{eq:eg-lin} however only applies to the case $K>M$. 

Linear networks thus show that the scaling of $\epsilon_g^*$ with $L$ is not only a
consequence of either specialisation or the mismatched range of the networks'
output functions, as one might have suspected by looking only at sigmoidal
networks.

Interestingly, if we rescale the learning rate by choosing $\eta=1/K$, we find 
that the generalisation error~\eqref{eq:eg-lin} becomes equal to $\sigma^2/2$, 
independent of $L$. Again, this rescaling of the learning rate comes at the cost 
of increased training time and hence, in this model, increased training data. 
A quantitative exploration of the trade-off between learning rate and network size 
for a fixed data set is an interesting problem, however, it goes beyond the domain 
of online learning and is hence left for future work.

\subsection{Numerical results suggest the same scalings for ReLU networks,
  too}\label{sec:excess-relu}

\begin{figure}[t]
  \centering
  \includegraphics[width=\linewidth]{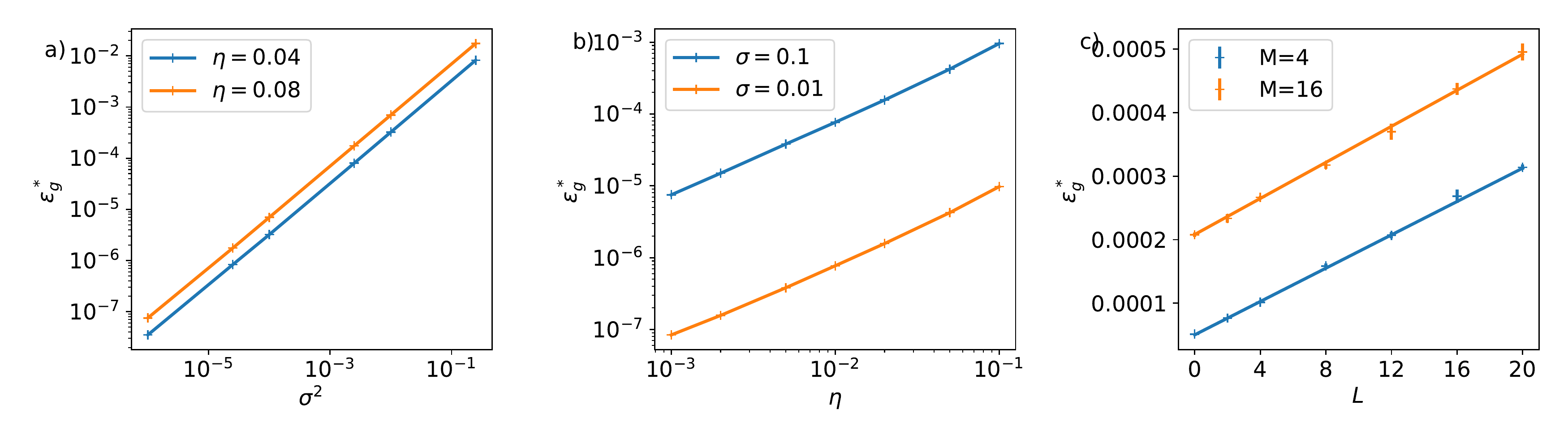}
  \caption{\label{fig:eg_relu_scaling} \textbf{The final generalisation error of
      over-parametrised ReLU networks scales as
      $\epsilon_g^* \sim \eta \sigma^2L$.} Simulations confirm that the
    asymptotic generalisation error $\epsilon_g^*$ of a ReLU student learning
    from a ReLU teacher scales with the learning rate $\eta$, the variance of
    the teacher's output noise $\sigma^2$ and the number of additional hidden
    units as $\epsilon_g\sim \eta \sigma^2L$, which is the same scaling as the
    one found analytically for sigmoidal networks in
    Eq.~\eqref{eq:egFinal1stOrderInLr}. Straight lines are linear fits to the
    data, with slope $1$ in (a) and (b). Parameters: $M=2, K=6$ (a, b) and
    $M=4, 16$; $K=M + L$ (c); in all figures, $N=784, \kappa=0$.}
\end{figure}

The analytical calculation of $\epsilon_g$ described above for networks with
ReLU activation function poses some additional technical challenges, so here we
resort to simulations to illustrate the behaviour of $\epsilon_g^*$ in this case
and leave the analytical description for future work. The results are shown in
Fig.~\ref{fig:eg_relu_scaling}: we found numerically that the asymptotic
generalisation error of a ReLU student learning from a ReLU teacher has the same
scaling as the one we found analytically for networks with sigmoidal and with
linear activation functions: $\epsilon_g^* \sim \eta \sigma^2 L$.

\subsubsection{Discussion}

Looking at the final overlap matrices $Q$ and
$R$ for ReLU networks in the right half of
Fig.~\ref{fig:final-representations-noisy}, we see a mechanism
behind Eq.~\eqref{eq:eg_scaling} for ReLU networks that is reminiscent of the linear case:
instead of the one-to-one specialisation of sigmoidal networks, all the hidden
units of the student have a finite overlap with all the hidden units of the
teacher. This is a consequence of the fact that it is much simpler to re-express
the sum of $M$ ReLU units with $K\neq M$ ReLU units. However, this also means
that there are a lot of redundant degrees of freedom, which nevertheless all
pick up fluctuations from the output noise and degrade the generalisation error.

For ReLU networks, one might imagine that several ReLU units can specialise to
one and only one teacher unit and thus act as an effective denoiser for that
teacher unit. However, we have checked numerically that this configuration is
not a stable fixed point of the SGD dynamics.

\section{Discussion}\label{sec:discussion}

The main result of the preceding section was a set of ODEs that described the
generalisation dynamics of over-parameterised two-layer neural networks. This
framework allowed us to derive the scaling of the generalisation error of the
student with the network size, the learning rate and the noise level in the
teacher's outputs. This scaling is robust with respect to the choice of the
activation function as it holds true for linear, sigmoidal and ReLU networks. In
this section, we discuss several possible changes to our setup and discuss their
impact on the scaling of $\epsilon_g^*$.

\subsection{Weight decay} A natural strategy to avoid overfitting is to
explicitly regularise the weights, for example by using weight decay. In our
setup, this is introduced by choosing a finite $\kappa>0$ in
Eq.~\eqref{eq:sgd}. In our simulations with $\kappa>0$, we did not find a
scenario where weight decay did not increase the final generalisation error
compared to the case~$\kappa=0$. In particular, we did not find a scenario where
weight decay improved the performance of a student with $L>0$. The corresponding
plots can be found in Appendix~\ref{sec:app-weight-decay}.

\subsection{SGD with mini-batches}

We also made sure that the phenomenology we observed persists if we move to
stochastic gradient descent with mini-batches, where the gradient estimate in
Eq.~\eqref{eq:sgd} is averaged over several samples $(x, y_B)$, as is standard
in practice. We observed that increasing the mini-batch size lowers the
asymptotic generalisation error up to a certain mini-batch size of order
$\sim100$ samples, after which it stays roughly constant. Crucially, having
mini-batches does not change the scaling of $\epsilon_g^*$ with $L$ (see
Appendix~\ref{sec:app-mini-batches} for details).

\subsection{Structured input data}

One idealised assumption in our setup is that we take our inputs $x$ as i.i.d.\
draws from a standard normal distribution (see Sec.~\ref{sec:setup}). We
therefore repeated our experiments using MNIST images as inputs $x$, while
leaving all other aspects of our setup the same. In particular, we still trained
the student on a regression task with $y$ generated using a random teacher. This
setup allowed us to trace any change in the generalisation behaviour of the
student to the higher-order correlations of the input distribution. Switching to
MNIST inputs reproduced the same $\epsilon_g$-$L$ curve as having Gaussian
inputs to within the experimental error; the interested reader is referred to
Appendix~\ref{sec:app-mnist} for a detailed description of these experiments.

\begin{figure}[t]
  \centering
  \includegraphics[width=\linewidth]{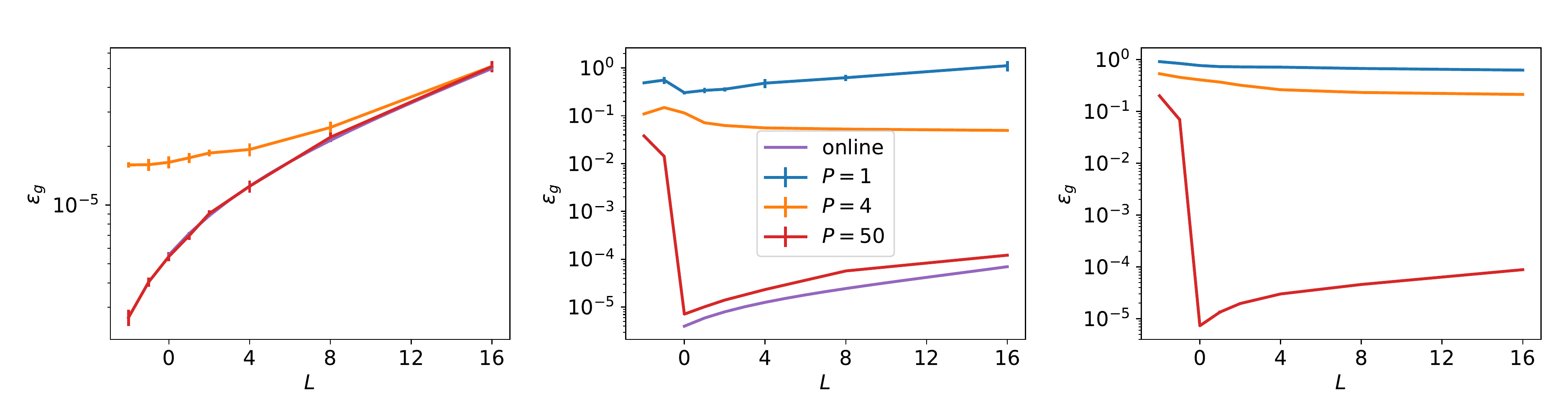}
  \caption{\label{fig:finite_ts} \textbf{The scaling of $\epsilon_g^*$ with $L$
      depends on the size of the training set.} We plot $\epsilon_g^*$ after SGD
    with a finite training set with $PN$ samples for linear, sigmoidal and ReLU
    networks from left to right. The result for online learning for linear and
    sigmoidal networks, Eqns.~\eqref{eq:egFinal} and~\eqref{eq:eg-lin}, are
    plotted in violet. In the linear case, the online learning result exactly
    matches the simulation for $P=50$. Error bars indicate one standard
    deviation over 10 simulations, each with a different training set; many of
    them are too small to be clearly visible. Parameters:
    $N=784, M=4, \eta=0.1, \sigma=0.01$.}
\end{figure}

\subsection{The scaling of $\epsilon_g^*$ with $L$ depends also on the size of the
  training set}

In practice, a single sample of the training data set will be
visited several times during training. After a first pass through the training
set, the online assumption that an incoming sample $(x, y_B)$ is uncorrelated to
the weights of the network thus breaks down. A complete analytical treatment in 
this setting remains an open problem, so to study this practically relevant
setup, we turn to simulations. We keep the setup described in Sec.~\ref{sec:setup},
but simply reduce the number of samples in the training data set $P$. Our focus
is again on the final generalisation error after convergence $\epsilon_g^*$ for
linear, sigmoidal and ReLU networks, which we plot from left to right 
as a function of $L$ in Fig.~\ref{fig:finite_ts}.

Linear networks show a similar behaviour to the setup with a very large training
set discussed in Sec.~\ref{sec:linear-networks}: the bigger the network, the
worse the performance for both $P=4$ and $P=50$. Again, the optimal network has
$K=1$ hidden units, irrespective of the size of the teacher. However, for
non-linear networks, the picture is more varied: For large training sets, where
the number of samples easily outnumber the free parameters in the network
($P=50$, red curve; this corresponds roughly to learning a data set of the size
of MNIST), the behaviour is qualitatively described by our theory from
Sec.~\ref{sec:final-eg}: the best generalisation is obtained by a network that
matches the teacher size, $K=M$. However, as we reduce the size of the training
set, this is no longer true. For $P=4$, for example, the best generalisation is
obtained with networks that have $K>M$. Thus the size of the training set with
respect to the network has an important influence on the scaling of
$\epsilon_g^*$ with $L$. Note that the early-stopping generalisation error,
which we define as the minimal generalisation error over the duration of
training, shows qualitatively the same behaviour as $\epsilon_g^*$ (see
Appendix~\ref{sec:app-early-stopping} for additional information.)

\section{Concluding perspectives}\label{sec:conclusion}

We have studied the dynamics of online learning in two-layer neural networks
within the teacher-student framework, where we train a student network using SGD on data
generated by another network, the teacher. One advantage of this setup is that
it allows us to investigate the behaviour of networks that are
over-parameterised with respect to the generative model of their data in a
controlled fashion. We derived a set of eight ODEs that describe the
generalisation dynamics of over-parameterised students of any size. Within this
framework, we analytically computed the final generalisation error of the student
in the limit of online learning with small noise. One immediate consequence of
this result is that SGD alone is not enough to regularise the over-parameterised
student, instead yielding networks whose generalisation error scales linearly
with the network size.

Furthermore, we demonstrated that adding explicit regularisation by introducing
weight decay did not improve the performance of the networks and that the same
phenomenology arises when using mini-batches or after substituting the Gaussian
inputs used in the theoretical analysis with a more realistic data
set. Nevertheless, we were able to find a scenario in our setup where the
generalisation decreases with the student's size, namely when training using a
finite data set that contains roughly as many samples as there are free
parameters in the network.

In the setting we analyse, our results clearly indicate that the regularisation 
of neural networks goes beyond the properties of SGD alone. Instead, a full
understanding of the generalisation properties of deep networks requires taking 
into account the interplay of at least the algorithm, its learning rate, the model
architecture, and the data set, setting up a formidable research programme 
for the future.

\section*{Acknowledgements}

SG and LZ acknowledge funding from the ERC under the European Union’s Horizon
2020 Research and Innovation Programme Grant Agreement 714608-SMiLe. MA thanks
the Swartz Program in Theoretical Neuroscience at Harvard University for
support. AS acknowledges funding by the European Research Council, grant 725937
NEUROABSTRACTION. FK acknowledges support from ``Chaire de recherche sur les
modèles et sciences des données'', Fondation CFM pour la Recherche-ENS, and from
the French National Research Agency (ANR) grant PAIL.

\clearpage

\appendix

\begin{center}
  \LARGE
  \textbf{APPENDICES}
\end{center}

\vspace*{1em}

\section{Derivation of the ODE description of the generalisation dynamics of
  online learning}
\label{sec:app-dynamics}

We will now show how to derive ODEs that describe the dynamics of online
learning in two-layer neural networks, following the seminal work by Biehl and
Schwarze~\cite{Biehl1995} and Saad and Solla~\cite{Saad1995a,Saad1995b}. We focus
on the teacher-student setup introduced in the main paper, where a student
network with weights $w\in\mathbb{R}^{K\times N}$ and output
\begin{equation}
  \phi(w, x) = \sum_{k=1}^K g\left(\frac{w_k x}{\sqrt{N}} \right)
\end{equation}
is trained on samples $(x^\mu, y^\mu)$ generated by another two-layer network
with weights $B\in\mathbb{R}^{M\times N}$, the teacher, according to
\begin{equation}
  \label{eq:app-y}
  y^\mu_B(x^\mu) \equiv \phi(B, x^\mu) + \zeta^\mu.
\end{equation}
Here, $\zeta^\mu$ is normally distributed with mean $0$ and variance $\sigma^2$.
We will make two technical assumptions, namely having a large network
($N\to\infty$) and a data set that is large enough to allow that we visit every
sample only once before training converges.

\subsection{Expressing the generalisation error in terms of order parameters}
\label{sec:app-expr-gener-error}

To make this section self-consistent, we briefly recapitulate how the assumptions
stated above allow to rewrite the generalisation error in terms of a number of
\emph{order parameters}. We have
\begin{align}
  \label{eq:app-eg}
  \epsilon_g \equiv & \frac{1}{2}\left\langle {\left[ \phi(w, x) - \phi(B,
                      x)\right]}^2 \right\rangle \\
  = & \frac{1}{2}\left\langle {\left[ \sum_{k=1}^K
      g\left(\lambda_k^\mu\right) - \sum_{m=1}^M g(\nu_m^\mu)\right]}^2
      \right\rangle, \label{eq:app-eg-explicit}
\end{align}
where we have introduced the local fields
\begin{align}
  \label{eq:app-local-fields}
  \lambda_k^\mu \equiv & \frac{w_k x^\mu}{\sqrt{N}}, \\
  \nu_m^\mu \equiv & \frac{B_m x^\mu}{\sqrt{N}}.
\end{align}
Here and throughout this paper, we will use the indices $i,j,k,\ldots$ to refer
to hidden units of the student, and indices $n,m,\ldots$ to denote hidden units
of the teacher. Since the input $x^\mu$ only appears in $\epsilon_g$ only via
products with the weights of the teacher and the student, we can replace the
high-dimensional average $\langle \cdot \rangle$ over the input distribution
$p(x)$ by an average over the $K+M$ local fields $\lambda_k^\mu$ and
$\nu_m^\mu$. The assumption that the training set is large enough to allow that
we visit every sample in the training set only once guarantees that the inputs
and the weights of the networks are uncorrelated. Taking the limit $N\to\infty$
ensures that the local fields are jointly normally distributed with mean zero
($\langle x_n \rangle=0$). Their covariance is also easily found: writing
$w_{ka}$ for the $a$th component of the $k$th weight vector, we have
\begin{equation}
  \label{eq:app-Q}
  \langle \lambda_k \lambda_l \rangle = \frac{\sum_{a,b}^N w_{ka} w_{lb}\langle
    x_a x_b \rangle}{N}=\frac{w_k w_l}{N} \equiv Q_{kl},
\end{equation}
since $\langle x_a x_b \rangle=\delta_{ab}$. Likewise, we define
\begin{equation}
  \label{eq:app-RandT}
  \langle \nu_n \nu_m \rangle = \frac{B_n B_m}{N} \equiv T_{nm}, \quad
  \langle \lambda_k \nu_m \rangle = \frac{w_k B_m}{N} \equiv R_{km}.
\end{equation}
The variables $R_{in}$, $Q_{ik}$, and $T_{nm}$ are called \emph{order
  parameters} in statistical physics and measure the overlap between student and
teacher weight vectors $w_i$ and $B_n$ and their self-overlaps,
respectively. Crucially, from Eq.~\eqref{eq:app-eg-explicit} we see that they are
sufficient to determine the generalisation error $\epsilon_g$. We can thus write
the generalisation error as
\begin{equation}
  \label{eq:app-eg_I2}
  \epsilon_g = \frac{1}{2} \sum_{i,k} I_2(i, k)
    + \frac{1}{2}\sum_{n, m} I_2(n, m)
    - \sum_{i, n} I_2 (i, n),
\end{equation}
where we have defined
\begin{equation}
  \label{eq:app-I2}
  I_2(i, k) \equiv \langle g(\lambda_i) g(\lambda_k) \rangle =  \frac{1}{\pi} \arcsin
  \frac{Q_{ik}}{\sqrt{1 + Q_{ii}}\sqrt{1 + Q_{kk}}}.
\end{equation}
The average in Eq.~\eqref{eq:app-I2} is taken over a normal distribution for the
local fields $\lambda_i$ and $\lambda_k$ with mean $(0, 0)$ and covariance
matrix
\begin{equation}
  C_2 = \begin{pmatrix}
    Q_{ii} & Q_{ik} \\
    Q_{ik} & Q_{kk}
  \end{pmatrix}.
\end{equation}
Since we are using the indices $i,j,\ldots$ for student units and $n,m,\ldots$
for teacher hidden units, we have
\begin{equation}
  I_2(i, n)=\langle g(\lambda_i)g(\nu_m) \rangle,
\end{equation}
where the covariance matrix of the joint of distribution $\lambda_i$ and $\nu_m$
is given by 
\begin{equation}
  C_2 = \begin{pmatrix}
    Q_{ii} & R_{in} \\
    T_{in} & T_{nn}
  \end{pmatrix}.
\end{equation}
and likewise for $I_2(n, m)$. We will use this convention to denote integrals
throughout this section. For the generalisation error, this means that it can be
expressed in terms of the order parameters alone as
\begin{multline}
  \label{eq:app-eg-order-parameters}
  \epsilon_g = \frac{1}{\pi} \sum_{i,k} \arcsin
  \frac{Q_{ik}}{\sqrt{1 + Q_{ii}}\sqrt{1 + Q_{kk}}}
  + \frac{1}{\pi} \sum_{n,m} \arcsin
  \frac{T_{nm}}{\sqrt{1 + T_{nn}}\sqrt{1 + T_{mm}}}\\
  - \frac{2}{\pi} \sum_{i,n} \arcsin
  \frac{R_{in}}{\sqrt{1 + Q_{ii}}\sqrt{1 + T_{nn}}}.
\end{multline}

\subsection{ODEs for the evolution of the order parameters}
\label{sec:app-odes-evolution-order}

Expressing the generalisation error in terms of the order parameters as we have
in Eq.~\eqref{eq:app-eg-order-parameters} is of course only useful if we can track
the evolution of the order parameters over time. We can derive ODEs that allow
us to do precisely that by first writing again the SGD update of the weights:
\begin{equation}
  \label{eq:app-sgd}
  w_k^{\mu+1} = w_k^{\mu} - \frac{\kappa}{N} w_k^\mu - \frac{\eta}{\sqrt{N}}
  x^\mu r_k^\mu,
\end{equation}
where $\mu$ is a running index counting the weight updates or, equivalently, the
samples used so far, and
\begin{equation}
  \label{eq:app-delta}
  r_k^\mu \equiv g'(\lambda_k^\mu) \left[ \phi(w, x^\mu)- y_B^\mu\right].
\end{equation}
From this equation, we can obtain differential equations for the time evolution
of the order parameters $Q$ by squaring the weight update~\eqref{eq:app-sgd} and for
$R$ taking the inner product of~\eqref{eq:app-sgd} with $B_n$, respectively, which
yields the Eqns.~(12) of the main text and which we state again for
completeness:
\begin{subequations}
  \label{eq:app-eom-short}
  \begin{align}
    \frac{\dd R_{in}}{\dd \alpha} &= -\kappa R_{in} + \eta \langle r_i \nu_n \rangle\\
    \frac{\dd Q_{ik}}{\dd \alpha} &= -2 \kappa Q_{ik} + \eta \langle r_i
      \lambda_k \rangle + \eta \langle r_k \lambda_i \rangle
      + \eta^2 \langle r_i r_k\rangle + \eta^2 \sigma^2 \langle
      g'(\lambda_i)g'(\lambda_k) \rangle
  \end{align}
\end{subequations}
where $\alpha=\mu / N$ becomes a continuous time-like variable in the limit
$N\to\infty$. These equations are valid for any choice of activation functions
$g_1$ and $g_2$. To make progress however, \emph{i.e.} to obtain a closed set of
differential equations for $Q$ and $R$, we need to evaluate the averages
$\langle \cdot \rangle$ over the local fields. In particular, we have to compute
three types of averages:
\begin{equation}
  I_3 = \langle g'(a) b g'(c) \rangle,
\end{equation}
where $a$ is one the local fields of the student, while $b$ and $c$ can be local
fields of either the student or the teacher;
\begin{equation}
  I_4 = \langle g'(a) g'(b) g(c) g(d) \rangle,
\end{equation}
where $a$ and $b$ are local fields of the student, while $c$ and $d$ can be
local fields of both; and finally
\begin{equation}
  \label{eq:app-2}
  J_2 = \langle g'(a)g'(b) \rangle,
\end{equation}
where $a$ and $b$ are local fields of the teacher. In each of these integrals,
the average is taken with respect to a multivariate normal distribution for the
local fields with zero mean and a covariance matrix whose entries are chosen in
the same way as discussed for $I_2$.

We can re-write Eqns.~\eqref{eq:app-eom-short} with these definitions in a more
explicit form as~\cite{Saad1995a,Saad1995b}
\begin{align}
  \label{eq:app-eom-long}
  \frac{\dd R_{in}}{\dd \alpha} & = -\kappa R_{in} + \eta \left( \sum_m I_3(i, n, m)
                                  - \sum_j I_3 (i, n, j)\right), \\
  \frac{\dd Q_{ik}}{\dd \alpha} & = -2\kappa Q_{ik} + \eta^2 \sigma^2 J_2(i, k)
                                  \nonumber \\
                                & \quad + \eta \left( \sum_m I_3(i, k, m) -
                                  \sum_j I_3(i, k, j) \right) \nonumber \\
                                & \quad + \eta \left( \sum_m I_3(k, i, m) - \sum_j I_3 (k, i, j) \right)
                                  \nonumber \\
                                & \quad + \eta^2 \left( \sum_{n,m} I_4(i, k, n, m) - 2 \sum_{j, n} I_4(i, k,
                                  j, n) + \sum_{j, l} I_4(i, k, j, l)\right).
\end{align}
The explicit form of the integrals $I_2(\cdot)$, $I_3(\cdot)$, $I_4(\cdot)$ and
$J_2(\cdot)$ is given in Sec.~\ref{sec:app-explicit-integrals} for the case
$g(x)=\erf\left(x/\sqrt{2}\right)$. Solving these equations numerically for $Q$
and $R$ and substituting their values in to the expression for the
generalisation error~\eqref{eq:app-eg_I2} gives the full generalisation dynamics
of the student. We show the resulting learning curves together with the result
of a single simulation in Fig.~2 of the main text. We have bundled our
simulation software and our ODE integrator as a user-friendly Python
package\footnote{To download, visit \url{https://github.com/sgoldt/pyscm}}. In
Sec.~\ref{sec:app-eg_analytical}, we discuss how to extract information from
them in an analytical way.

\section{Calculation of $\epsilon_g$ in the limit of small noise}
\label{sec:app-eg_analytical}

Our aim is to understand the asymptotic value of the generalisation error
\begin{equation}
   \epsilon_g^* \equiv \lim_{\alpha\to\infty} \epsilon_g(\alpha).
\end{equation}
We focus on students that have more hidden units than the teacher, $K\ge
M$. These students are thus over-parameterised \emph{with respect to the
  generative model of the data} and we define
\begin{equation}
  \label{eq:app-L}
  L \equiv K  - M
\end{equation}
as the number of additional hidden units in the student network. In this
section, we focus on the sigmoidal activation function
\begin{equation}
  \label{eq:app-erf}
  g(x) = \erf\left(x/\sqrt{2}\right),
\end{equation}
unless stated otherwise.

Eqns.~\eqref{eq:app-eom-long} are a useful tool to analyse the generalisation
dynamics and they allowed Saad and Solla to gain plenty of analytical insight
into the special case $K=M$~\cite{Saad1995a,Saad1995b}. However, they are also a
bit unwieldy. In particular, the number of ODEs that we need to solve grows with
$K$ and $M$ as $K^2+K M$. To gain some analytical insight, we make use of the
symmetries in the problem, \emph{e.g.}  the permutation symmetry of the hidden
units of the student, and re-parametrised the matrices $Q_{ik}$ and $R_{in}$ in
terms of eight order parameters that obey a set of self-consistent ODEs for any
$K>M$. We choose the following parameterisation with eight order parameters:
\begin{align}
  \label{eq:app-ansatz}
  Q_{ij} =& \begin{cases}
    Q & \quad i=j \le M, \\
    C & \quad i \neq j; \; i,j \le M,\\
    D & \quad i > M, j \le M \quad \mathrm{or}\quad i \le M, j > M,\\
    E & \quad i=j > M,\\
    F & \quad i \neq j; \; i,j > M,
  \end{cases}\\
  R_{in} =& \begin{cases}
    R & \quad i=n, \\
    S & \quad i \neq n; \; i \le M,\\
    U & \quad i > M,
  \end{cases}
\end{align}
which in matrix form for the case $M=3$ and $K=5$ read:
\begin{equation}
  R = \begin{pmatrix}
    R & S & S \\
    S & R & S \\
    S & S & R \\
    U & U & U \\
    U & U & U
  \end{pmatrix} \quad \mathrm{and} \quad 
  Q = \begin{pmatrix}
    Q & C & C & D & D \\
    C & Q & C & D & D \\
    C & C & Q & D & D \\
    D & D & D & E & F \\
    D & D & D & F & E \\
  \end{pmatrix}
\end{equation}
We choose this number of order parameters and this particular setup for the
overlap matrices $Q$ and $R$ for two reasons: it is the smallest number of
variables for which we were able to self-consistently close the equations of
motion~\eqref{eq:app-eom-long}, and they agree with numerical evidence obtained from
integrating the full equations of motion~\eqref{eq:app-eom-long}.

By substituting this ansatz into the equations of
motion~\eqref{eq:app-eom-long}, we find a set of eight ODEs for the order
parameters. These equations are rather unwieldy and some of them do not even fit
on one page, which is why we do not print them here in full; instead, we provide
a \emph{Mathematica} notebook where they can be found and interacted
with\footnote{To download, visit \url{https://github.com/sgoldt/pyscm}}. These equations
allow for a detailed analysis of the effect of over-parameterisation on the
asymptotic performance of the student, as we will discuss now.

\subsection{Heavily over-parameterised students can learn perfectly from a
  noiseless teacher using online learning }
\label{sec:app-no-noise}

\begin{figure}
  \centering
  \includegraphics[width=\textwidth]{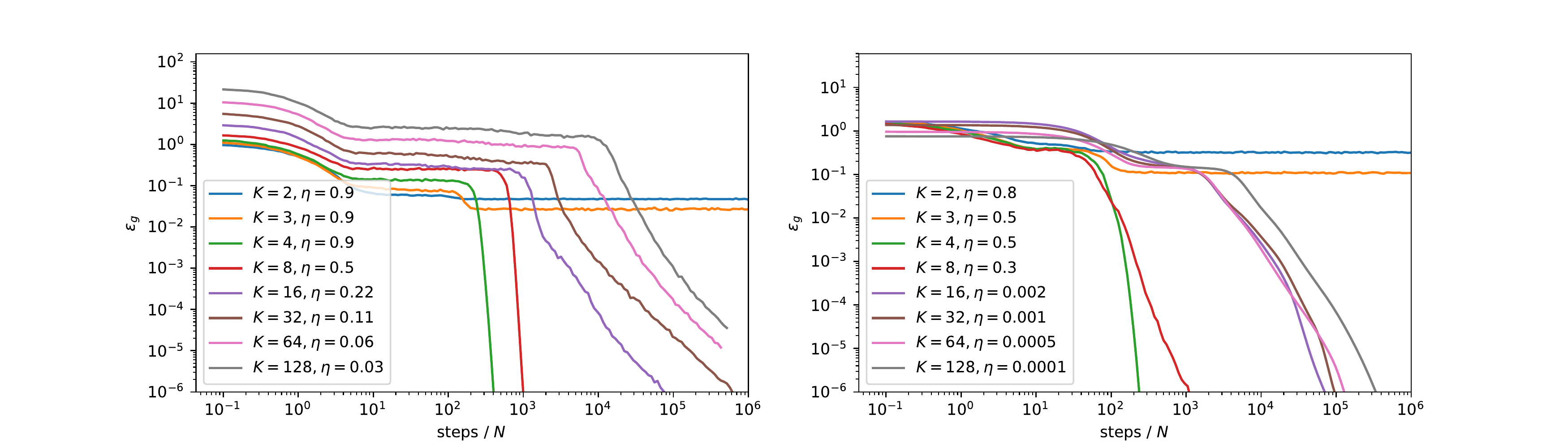}%
  \caption{\label{fig:app-eg-sigma0} \textbf{Over-parametrised networks with
      sigmoidal or ReLU activations learn perfectly from a noiseless teacher.}
    The generalisation dynamics for students with sigmoidal (left) and ReLU
    activation function (right) for various $K$ learning from a teacher with
    $M = 4$ is shown. In all cases, the generalisation error eventually decays
    exponentially towards zero. ($N=784$)}
\end{figure}

For a teacher with $T_{nm}=\delta_{nm}$ and in the absence of noise in the
teacher's outputs ($\sigma=0$), there exists a fixed point of the ODEs with
$R=Q=1$, $C=D=E=F=0$, and perfect generalisation $\epsilon_g=0$.  Online
learning will find this fixed point, as is demonstrated in
Fig.~\ref{fig:app-eg-sigma0}, where we plot the generalisation dynamics of a student
with $K$ hidden units learning from a teacher with $M=4$ hidden units for both
Erf and ReLU activation functions. More precisely, after a plateau whose length
depends on the size of the network for the sigmoidal network, the generalisation
error eventually begins an exponential decay to the optimal solution with zero
generalisation error. The learning rates are chosen such that learning
converges, but aren't optimised otherwise.

\subsection{Perturbative solution of the ODEs}

We have calculated the asymptotic value of the generalisation error
$\epsilon_g^*$ for a teacher with $T_{nm}=\delta_{nm}$ to first order in the
variance of the noise $\sigma^2$. To do so, we performed a perturbative expansion
around the fixed point
\begin{gather}
  R_0=Q_0=1,\\
  S_0=U_0=C_0=D_0=E_0=F_0=0,
\end{gather}
with the ansatz
\begin{equation}
  X = X_0 + \sigma^2 X_1
\end{equation}
for all the order parameters. Writing the ODEs to first order $\sigma^2$ and
solving for their steady state where $X'(\alpha)=0$ yielded a fixed point with
an asymptotic generalisation error
\begin{equation}
  \label{eq:app-egFinal}
  \epsilon_g^* = \frac{\sigma^2 \eta}{2 \pi} f(M, L, \eta) + \mathcal{O}(\sigma^3).
\end{equation}
$f(M, L, \eta)$ is an unwieldy rational function of its variables. Due to its
length, we do not print it here in full; instead, we give the full function in a
\emph{Mathematica} notebook\footnote{To download, visit
  \url{https://github.com/sgoldt/pyscm}}. Here, we plot the results in various forms in
Fig.~\ref{fig:app-eg_erf_scaling}. We note in particular the following points:

\begin{figure}
  \centering
  \includegraphics[width=\linewidth]{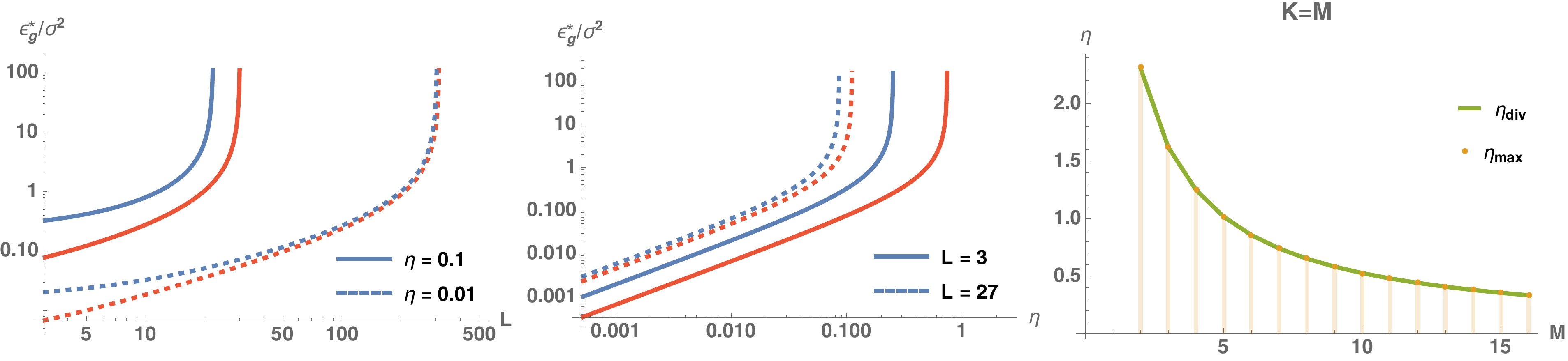}
  \caption{\label{fig:app-eg_erf_scaling} \textbf{The final generalisation error of
      over-parameterised Erf networks scales linearly with the learning rate,
      the variance of the teacher's output noise, and $L$.} We plot
    $\epsilon_g^*/\sigma^2$ in the limit of small noise, Eq.~\eqref{eq:app-egFinal},
    for $M=2$ (red) and $M=16$ (blue). It is clear that generalisation error
    increases with the number of superfluous units $L$ at fixed learning rate
    (\emph{left}) and the learning rate $\eta$ (\emph{middle}). \emph{Right:}
    For $K=M$, the learning rate $\eta_{\mathrm{div}}$ at which our perturbative
    result diverges is precisely the maximum learning rate $\eta_{\max}$
    at which the exponential convergence to the optimal solution is guaranteed
    for $\sigma=0$, Eq.~\eqref{eq:app-etaMax}}
\end{figure}

\subsubsection{Discussion}
\label{sec:app-eg-discussion}

\begin{figure}
  \centering
  \includegraphics[width=.5\linewidth]{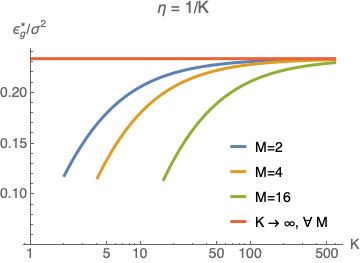}
  \caption{\label{fig:app-eg_erf_lr_rescaled} \textbf{Asymptotic generalisation
      error for sigmoidal networks with learning rate $\eta=1/K$.} We plot the
    asymptotic generalisation error $\epsilon_g^*$~\eqref{eq:app-egFinal} over
    $\sigma^2$ of a student with a varying number of hidden units trained on
    data generated by teachers with $M=2, 4, 16$ using SGD with learning rate
    $1/K$. The generalisation error still increases with $K$, before plateauing
    at a constant value that is independent of $M$. $\kappa=0$.}
\end{figure}

\begin{description}
\item[$\epsilon_g^*$ increases with $L$, $\eta$] The two plots on the left show
  that the generalisation error increases monotonically with both $L$ and $\eta$
  while keeping the other fixed, resp., for teachers with $M=2$ (red) and $M=16$
  (blue)
\item[Divergence at large $\eta$] Our perturbative result diverges for large
  $L$, or equivalently, for a large learning rate that depends on the number of
  hidden units $L\sim K$. For the special case $K=M$, the learning rate
  $\eta_{\mathrm{div}}$ at which our perturbative result diverges is precisely
  the maximum learning rate $\eta_{\mathrm{max}}$ for which the exponential
  convergence to the optimal solution is still guaranteed for
  $\sigma=0$~\cite{Saad1995b}
  \begin{equation}
  \label{eq:app-etaMax}
  \eta_{\max} = \frac{\sqrt{3} \pi }{M+3/\sqrt{5}-1}
\end{equation}
  as we show in the right-most plot of Fig.~\ref{fig:app-eg_erf_scaling}.
\item[Expansion for small $\eta$] In the limit of small learning rates, which is
  the most relevant in practice and which from the plots in
  Fig.~\ref{fig:app-eg_erf_scaling} dominates the behaviour of $\epsilon_g^*$
  outside of the divergence, the generalisation error is linear in the learning
  rate. Expanding $\epsilon_g^*$ to first order in the learning rate reveals a
  particularly revealing form,
  \begin{equation}
    \label{eq:app-egFinal1stOrderInLr}
    \epsilon_g^* = \frac{\sigma^2 \eta}{2 \pi} \left(L + \frac{M}{\sqrt{3}} \right) + \mathcal{O}(\eta^2)
  \end{equation}
  with second-order corrections that are quadratic in $L$. This is actually the
  sum of the asymptotic generalisation errors of $M$ continuous perceptrons that
  are learning from a teacher with $T=1$ and $L$ continuous perceptrons with
  $T=0$ as we calculate in Sec.~\ref{sec:app-cp}. This neat result is a
  consequence of the specialisation that is typical of SCMs with sigmoidal
  activation functions as we discussed in the main text.
\item[Rescaling the learning rate by $K$] The expression for the generalisation
  error in the limit of small learning rates might tempt one to rescale the
  learning rate $\eta$ by $K$ in order to mitigate the detrimental effect of the
  over-parameterisation. As we note in the main text, this a leads to a longer
  training duration which in our model implies that more data is required until
  the final generalisation error is achieved, both of which might not be
  feasible in practice. Moreover, we show in Fig.~\ref{fig:app-eg_erf_lr_rescaled}
  that the asymptotic generalisation error~\eqref{eq:app-egFinal} of a student
  trained using SGD with learning rate $\eta=1/K$ still increases with $L$
  before plateauing at a constant value that is independent of $M$.
\end{description}

\section{Asymptotic generalisation error of a noisy continuous perceptron}
\label{sec:app-cp}

What is the asymptotic generalisation for a continuous perceptron, \emph{i.e.} a
network with $K=1$, in a teacher-student scenario when the teacher has some
additive Gaussian output noise? In this section, we repeat a calculation
by Biehl and Schwarze~\cite{Biehl1995} where the teacher's outputs are given by
\begin{equation}
  y_B = g \left(\frac{B x}{\sqrt{N}}\right) + \zeta
\end{equation}
where $\zeta$ is again a Gaussian random variable with mean 0 and variance
$\sigma^2$. We keep denoting the weights of the student by $w$ and the weights
of the teacher by $B$. To analyse the generalisation dynamics, we introduce the
order parameters
\begin{equation}
   R \equiv \frac{w B}{N}, \qquad Q \equiv \frac{w w}{N} \quad\mathrm{and}\quad T \equiv \frac{B B}{N}.
\end{equation}
and we explicitly do not fix $T$ for the moment. For $g(x)=\mathrm{erf}\left(x/\sqrt{2}\right)$,
they obey the following equations of motion:
\begin{align}
  \frac{d R}{d t} =&\frac{2 \eta}{\pi  \left(Q(t)+1\right)}  \left(\frac{T Q(t)-R(t)^2+T}{\sqrt{(T+1) Q(t)-R(t)^2+T+1}}-\frac{R(t)}{\sqrt{2 Q(t)+1}}\right) \\
  \frac{d Q}{d t} =& \frac{4 \eta}{\pi  (Q(t)+1)}  \left(\frac{R(t)}{\sqrt{2 (Q(t)+1)-R(t)^2}}-\frac{Q(t)}{\sqrt{2
                     Q(t)+1}}\right)\nonumber \\
                   & + \frac{4 \eta ^2}{\pi ^2 \sqrt{2 Q(t)+1}} \left[ -2 \arcsin\left(\frac{R(t)}{ \sqrt{(6 Q(t)+2)(2 Q(t)-R(t)^2+1)}}\right) \right. \nonumber \\
                   & \qquad +\left. \arcsin\left(\frac{2
                     \left(Q(t)-R(t)^2\right)+1}{2 \left(2 Q(t)-R(t)^2+1\right)}\right)+\arcsin
                     \left(\frac{Q(t)}{3 Q(t)+1}\right)\right]\nonumber \\
                   & +\frac{2 \eta ^2 \sigma ^2}{\pi  \sqrt{2 Q(t)+1}}.
\end{align}

The equations of motion have a fixed point at $Q=R=T$ which has perfect
generalisation for $\sigma=0$. We hence make a perturbative ansatz in $\sigma^2$
\begin{align}
    Q(t) =& T + \sigma^2 q(t) \\
    R(t) =& T + \sigma^2 r(t)
\end{align}
and find for the asymptotic generalisation error
\begin{equation}
    \epsilon_g^* = \frac{\eta  \sigma ^2 (4 T+1)}{2 \sqrt{2 T+1} \left(-\eta  \sqrt{8 T^2+6 T+1}+4
   \pi  T+\pi \right)} + \mathcal{O}\left(\sigma^3\right).
\end{equation}
To first order in the learning rate, this reads
\begin{equation}
    \epsilon_g^* = \frac{\eta  \sigma ^2}{2 \pi  \sqrt{2 T+1}},
\end{equation}
which should be compared to the corresponding result for the full SCMs,
Eq.~\eqref{eq:app-egFinal1stOrderInLr}.

\section{Regularisation by weight decay does not help}
\label{sec:app-weight-decay}

\begin{figure}
  \centering
  \includegraphics[width=\linewidth]{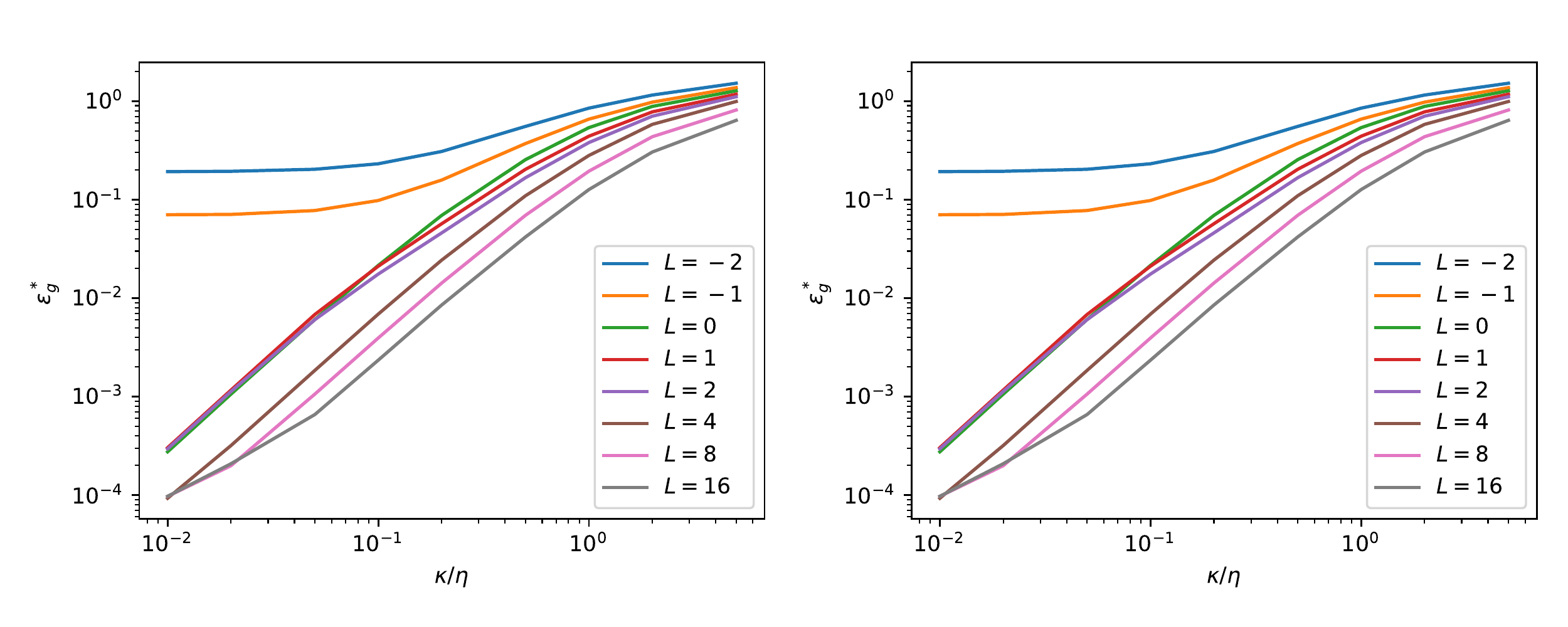}
  \caption{\label{fig:app-weight-decay} \textbf{Weight decay.} We plot the final
    generalisation error $\epsilon_g^*$ of a student with a varying number of
    hidden units trained on data generated by a teacher with $M=4$ using SGD
    with weight decay. The generalisation error clearly increases with the
    weight decay constant $\kappa$. Parameters: $N=784, \eta=0.1, sigma=0.01$.}
\end{figure}

A natural strategy to avoid the pitfalls of overfitting is to regularise the
weights, for example by using explicit weight decay by choosing $\kappa>0$. We
have not found a setup where adding weight decay \emph{improved} the asymptotic
generalisation error of a student compared to a student that was trained without
weight decay in our setup. As a consequence, weight decay completely fails to
mitigate the increase of $\epsilon_g^*$ with $L$. We show the results of an
illustrative experiment in Fig.~\ref{fig:app-weight-decay}.

\section{SGD with mini-batches}
\label{sec:app-mini-batches}

\begin{figure}
  \centering
  \includegraphics[width=\textwidth]{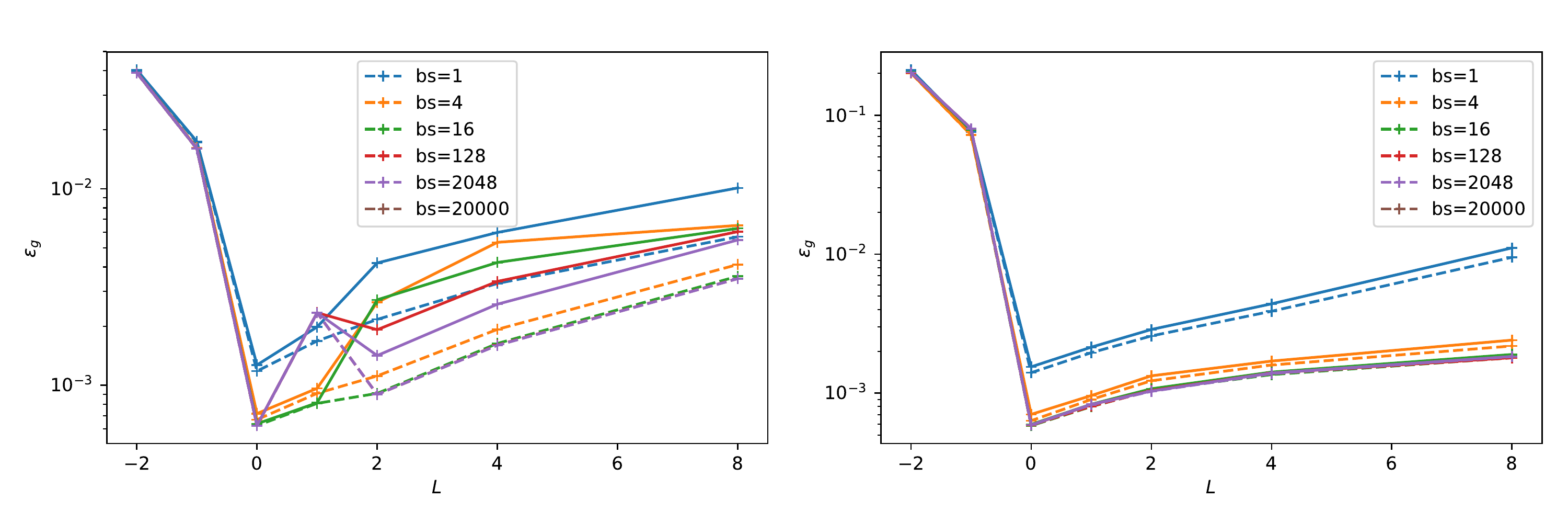}%
  \caption{\label{fig:app-minibatches} \textbf{SGD with mini-batches shows the same
      qualitative behaviour as online learning} We show the asymptotic
    generalisation error $\epsilon_g^*$ for students with sigmoidal (left) and
    ReLU activation function (right) for various $K$ learning from a teacher
    with $M = 4$. Between the curves, we change the size of the mini-batch used
    at each step of SGD from $1$ (online learning) to 20 000. Parameters:
    $N=500, \eta=0.2, \sigma=0.1, \kappa=0$.}
\end{figure}

One key characteristic of online learning is that we evaluate the gradient of
the loss function using a single sample from the training step per step. In
practice, it is more common to actually use a number of samples $b>1$ to
estimate the gradient at every step. To be more precise, the weight
update equation for SGD with mini-batches would read:
\begin{equation}
  \label{eq:app-sgd-mb}
  w_k^{\mu+1} = w_k^{\mu} - \frac{\kappa}{N} w_k^\mu - \frac{\eta}{b\sqrt{N}}
  \sum_{\ell=1}^b x^{\mu,\ell} g'(\lambda_k^{\mu,\ell}) \left[ \phi(w, x^{\mu,\ell})- y_B^{\mu,\ell}\right].
\end{equation}
where $x^{\mu,\ell}$ is the $\ell$th input from the mini-batch used in the $m$th
step of SGD, $\lambda_k^{\mu,\ell}$ is the local field of the $k$th student unit
for the $\ell$th sample in the mini-batch, etc. Note that when we use every
sample only once during training, using mini-batches of size $b$ increases the
amount of data required by a factor $b$ when keeping the number of steps
constant.

We show the asymptotic generalisation error of student networks of varying size
trained using SGD with mini-batches and a teacher with $M=4$ in
Fig.~\ref{fig:app-minibatches}. Two trends are visible: first, using increasing the
size of the mini-batches decreases the asymptotic generalisation error
$\epsilon_g^*$ up to a certain mini-batch size, after which the gains in
generalisation error become minimal; and second, the shape of the
$\epsilon_g^*-L$ curve is the same for all mini-batch sizes, with the minimal
generalisation error attained by a network with $K=M$.

\section{Using MNIST images for training and testing}
\label{sec:app-mnist}

\begin{figure}
  \centering
  \includegraphics[width=\linewidth]{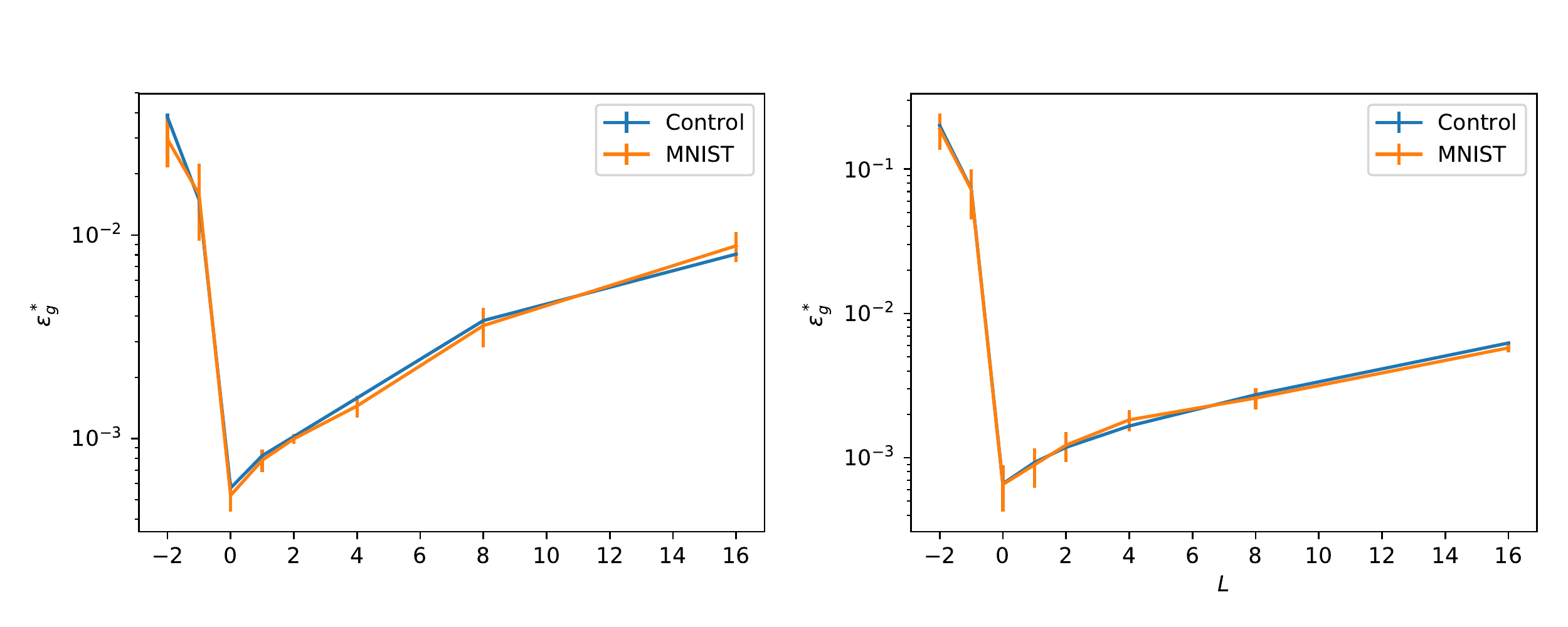}
  \caption{\label{fig:app-mnist-inputs} \textbf{Higher-order correlations in the
      input data do not play a role for the asymptotic generalisation.} WE plot
    the final generalisation error $\epsilon_g^*$ after online learning of a
    student of various sizes from a teacher with $M=4$ using Gaussian inputs
    (blue) and MNIST images (red) for training and testing.
    $N=784, \eta=0.1, \sigma=0.1, \kappa=0$.}
\end{figure}

In the derivation of the ODE description of online learning for the main text,
we noted that only the first two moments of the input distribution matter for
the learning dynamics and for the final generalisation error. The reason for
this is that the inputs only appear in the equations of motion for the order
parameters as a product with the weights of either the teacher or the
student. Now since they are -- by assumption -- uncorrelated with those weights,
this product is the sum of large number of random variables and hence
distributed by the central limit theorem.

We have checked how our results change when this assumption breaks down in one
example where we train a network on a finite data set with non-trivial higher
order moments, namely the images of the MNIST data set. We studied the very same
setup that we discuss throughout this work, namely the supervised learning of a
regression task in the teacher-student scenario. We \emph{only} replace the the
inputs, which would have been i.i.d.\ draws from the standard normal
distribution, with the images of the MNIST data set. In particular, this means
that we do not care about the labels of the
images. Figure~\ref{fig:app-mnist-inputs} shows a plot of the resulting final
generalisation against $L$ for both the MNIST data set and a data set of the same
size, comprised of i.i.d.\ draws from the standard normal distribution, which
are in good agreement.

\section{Early-stopping generalisation error for finite training sets}
\label{sec:app-early-stopping}

\begin{figure}
  \centering
  \includegraphics[width=\linewidth]{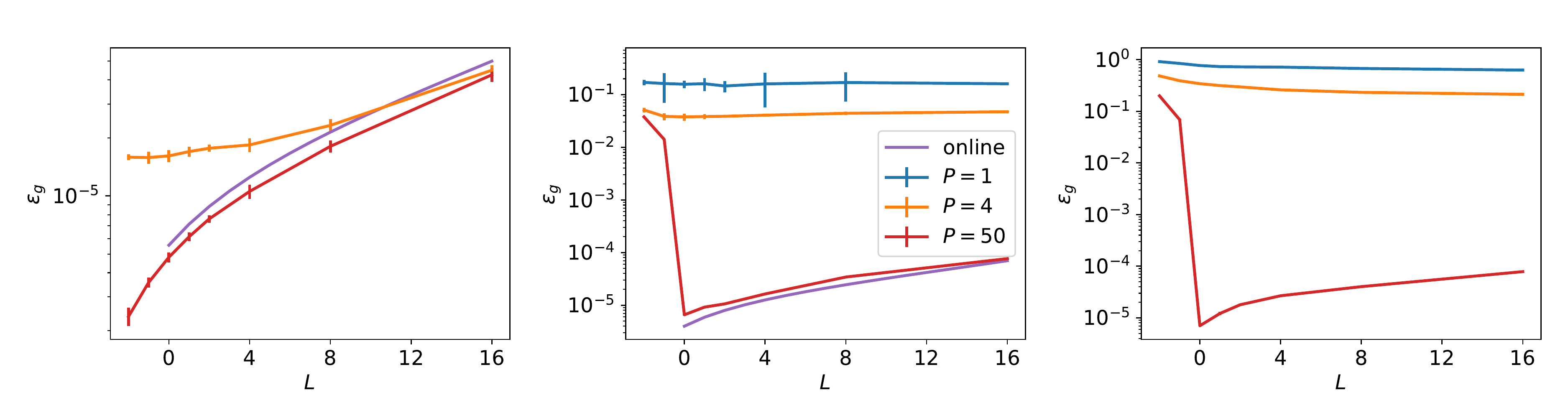}
  \includegraphics[width=\linewidth]{finite_ts_eg_final}
  \caption{\label{fig:app-finite-ts-eg-early}\textbf{The scaling of $\epsilon_g^*$
      with $L$ shows a similar dependence on the size of the training set for
      early-stopping (top) and final (bottom) generalisation error.} We plot the
    asymptotic and the early-stopping generalisation error after SGD with a
    finite training set containing $PN$ samples (linear, sigmoidal and ReLU
    networks from left to right). The result for online learning for linear and
    sigmoidal networks, Eqns.~13 and~15 of the main text, are plotted in
    violet. Error bars indicate one standard deviation over 10 simulations, each
    with a different training set; many of them are too small to be clearly
    visible. Parameters: $N=784, M=4, \eta=0.1, \sigma=0.01$.}
\end{figure}

A common way to prevent over-fitting of a neural network when training with a
finite training set in practice is early stopping, where the training is stopped
before the training error has converged to its final value yet. The idea behind
early-stopping is thus to stop training before over-fitting sets in. For the
purpose of our analysis of the generalisation of two-layer networks trained on a
fixed finite data set in Sec.~4 of the main text, we define the early-stopping
generalisation error $\hat{\epsilon}_g$ as the minimum of $\epsilon_g$ during
the whole training process. In Fig.~\ref{fig:app-finite-ts-eg-early}, we
reproduce Fig.~6 from the main text at the bottom and plot $\hat{\epsilon}_g$
obtained from the very same experiments at the top. While the ReLU networks
showed very little to no over-training, the sigmoidal networks showed more
significant over-training. However, the qualitative dependence of the
generalisation errors on $L$ was observed to be the same in this experiment. In
particular, the early-stopping generalisation error also shows two different
regimes, one where increasing the network hurts generalisation ($P\gg K$), and
one where it improves generalisation or at least doesn't seem to affect it much
(small $P\sim K$).

\section{Explicit form of the integrals appearing in the equations of motion of
  sigmoidal networks}
\label{sec:app-explicit-integrals}

To be as self-contained as possible, here we collect the explicit forms of the
integrals $I_2$, $I_3$, $I_4$ and $J_2$ that appear in the equations of motion
for the order parameters and the generalisation error for networks with
$g(x)=\erf\left( x/\sqrt{2} \right)$, see Eq.~\eqref{eq:app-eom-long}. They were
first given by~\cite{Biehl1995,Saad1995a}. Each average $\langle \cdot \rangle$
is taken w.r.t.\ a multivariate normal distribution with mean 0 and covariance
matrix $C\in\mathbb{R}^n$, whose components we denote with small letters. The
integration variables $u, v$ are always components of $\lambda$, while $w$ and
$z$ can be components of either $\lambda$ or $\nu$.
\begin{align}
  J_2 & \equiv \langle  g'(u) g'(v) \rangle =  \frac{2}{\pi}{\left( 1 + c_{11} +
        c_{22} + c_{11}c_{22}  - c_{12}^2       \right)}^{-1/2} \\[0.5em]
  I_2 & \equiv \frac{1}{2}\langle g(w)g(z) \rangle =  \frac{1}{\pi} \arcsin
        \frac{c_{12}}{\sqrt{1 + c_{11}}\sqrt{1 + c_{12}}}.\\[.5em]
  I_3 & \equiv \langle g'(u) w g(z) \rangle =
        \frac{2}{\pi}\frac{1}{\sqrt{\Lambda_3}} \frac{c_{23}(1 + c_{11}) -
        c_{12}c_{13}}{1 + c_{11}}\\[.5em]
  I_4 & \equiv \langle g'(u) g'(v) g(w) g(z) \rangle = \frac{4}{\pi^2}\frac{1}{\sqrt{\Lambda_4}}\arcsin\left( \frac{\Lambda_0}{\sqrt{\Lambda_1\Lambda2}} \right)
\end{align}
where
\begin{equation}
  \Lambda_4 = (1 + c_{11})  (1 + c_{22}) - c_{12}^2
\end{equation}
and
\begin{align}
  \Lambda_0 &= \Lambda_4 c_{34} - c_{23}c_{24}(1 + c_{11}) - c_{13}c_{14}(1 +
  c_{22}) + c_{12}c_{13}c_{24} + c_{12}c_{14}c_{23} \\[0.5em]
  \Lambda_1 &= \Lambda_4 (1 + c_{33}) - c_{23}^2(1 + c_{11}) - c_{13}^2(1 +
                c_{22}) + 2 c_{12}c_{13}c_{23} \\[0.5em]
  \Lambda_2 & = \Lambda_4 (1 + c_{44}) - c_{24}^2(1 + c_{11}) - c_{14}^2(1 +
                c_{22}) + 2 c_{12}c_{14}c_{24}
\end{align}

\bibliographystyle{unsrt}
\bibliography{scm}

\begin{thebibliography}{10}

\bibitem{Lecun2015}
Y.~LeCun, Y.~Bengio, and G.~E. Hinton.
\newblock {Deep learning}.
\newblock {\em Nature}, 521(7553):436--444, 2015.

\bibitem{Mnih2015}
V.~Mnih, K.~Kavukcuoglu, D.~Silver, A.A. Rusu, J.~Veness, M.G. Bellemare,
  A.~Graves, M.~Riedmiller, A.K. Fidjeland, G.~Ostrovski, S.~Petersen,
  C.~Beattie, A.~Sadik, I.~Antonoglou, H.~King, D.~Kumaran, D.~Wierstra,
  S.~Legg, and D.~Hassabis.
\newblock {Human-level control through deep reinforcement learning}.
\newblock {\em Nature}, 518(7540):529--533, 2015.

\bibitem{Silver2016}
D.~Silver, A.~Huang, C.J. Maddison, A.~Guez, L.~Sifre, G.~van~den Driessche,
  J.~Schrittwieser, I.~Antonoglou, V.~Panneershelvam, M.~Lanctot, S.~Dieleman,
  D.~Grewe, J.~Nham, N.~Kalchbrenner, I.~Sutskever, T.~Lillicrap, M.~Leach,
  K.~Kavukcuoglu, T.~Graepel, and D.~Hassabis.
\newblock {Mastering the game of Go with deep neural networks and tree search}.
\newblock {\em Nature}, 529(7587):484--489, 2016.

\bibitem{Simonyan2015}
K.~Simonyan and A.~Zisserman.
\newblock {Very Deep Convolutional Networks for Large-Scale Image Recognition}.
\newblock In {\em International Conference on Learning Representations}, 2015.

\bibitem{Bartlett2003}
P.L. Bartlett and S.~Mendelson.
\newblock {Rademacher and Gaussian complexities: Risk bounds and structural
  results}.
\newblock {\em Journal of Machine Learning Research}, 3(3):463--482, 2003.

\bibitem{Mohri2012}
Mehryar Mohri, Afshin Rostamizadeh, and Ameet Talwalkar.
\newblock {\em {Foundations of Machine Learning}}.
\newblock MIT Press, 2012.

\bibitem{Neyshabur2015a}
B.~Neyshabur, R.~Tomioka, and N.~Srebro.
\newblock {Norm-Based Capacity Control in Neural Networks}.
\newblock In {\em Conference on Learning Theory}, 2015.

\bibitem{Golowich2017}
N.~Golowich, A.~Rakhlin, and O.~Shamir.
\newblock {Size-Independent Sample Complexity of Neural Networks}.
\newblock {\em arxiv:1712.06541}, 2017.

\bibitem{Dziugaite2017}
G.K. Dziugaite and D.M. Roy.
\newblock {Computing Nonvacuous Generalization Bounds for Deep (Stochastic)
  Neural Networks with Many More Parameters than Training Data}.
\newblock In {\em Proceedings of the Thirty-Third Conference on Uncertainty in
  Artificial Intelligence}, 2017.

\bibitem{Arora2018}
S.~Arora, R.~Ge, B.~Neyshabur, and Yi~Zhang.
\newblock {Stronger generalization bounds for deep nets via a compression
  approach}.
\newblock {\em arxiv:1802.05296}, 2018.

\bibitem{Allen-Zhu2018}
Z.~Allen-Zhu, Y.~Li, and Y.~Liang.
\newblock {Learning and Generalization in Overparameterized Neural Networks,
  Going Beyond Two Layers}.
\newblock {\em arXiv:1811.04918}, 2018.

\bibitem{Neyshabur2015}
B.~Neyshabur, R.~Tomioka, and N.~Srebro.
\newblock {In search of the real inductive bias: On the role of implicit
  regularization in deep learning}.
\newblock In {\em ICLR}, 2015.

\bibitem{Zhang2016a}
C.~Zhang, S.~Bengio, M.~Hardt, B.~Recht, and O.~Vinyals.
\newblock {Understanding deep learning requires rethinking generalization}.
\newblock In {\em ICLR}, 2017.

\bibitem{Arpit2017}
D.~Arpit, S.~Jastrz, M.~S. Kanwal, T.~Maharaj, A.~Fischer, A.~Courville, and
  Y.~Bengio.
\newblock {A Closer Look at Memorization in Deep Networks}.
\newblock In {\em Proceedings of the 34th International Conference on Machine
  Learning}, 2017.

\bibitem{Biehl1995}
M.~Biehl and H.~Schwarze.
\newblock {Learning by on-line gradient descent}.
\newblock {\em J. Phys. A. Math. Gen.}, 28(3):643--656, 1995.

\bibitem{Saad1995a}
D.~Saad and S.A. Solla.
\newblock {Exact Solution for On-Line Learning in Multilayer Neural Networks}.
\newblock {\em Phys. Rev. Lett.}, 74(21):4337--4340, 1995.

\bibitem{Saad1995b}
D.~Saad and S.A. Solla.
\newblock {On-line learning in soft committee machines}.
\newblock {\em Phys. Rev. E}, 52(4):4225--4243, 1995.

\bibitem{Gardner1989}
E.~Gardner and B.~Derrida.
\newblock {Three unfinished works on the optimal storage capacity of networks}.
\newblock {\em Journal of Physics A: Mathematical and General},
  22(12):1983--1994, 1989.

\bibitem{Seung1992}
H.~S. Seung, H.~Sompolinsky, and N.~Tishby.
\newblock {Statistical mechanics of learning from examples}.
\newblock {\em Physical Review A}, 45(8):6056--6091, 1992.

\bibitem{Watkin1993}
T.~L.~H. Watkin, A.~Rau, and M.~Biehl.
\newblock {The statistical mechanics of learning a rule}.
\newblock {\em Reviews of Modern Physics}, 65(2):499--556, 1993.

\bibitem{Engel2001}
A.~Engel and C.~{Van den Broeck}.
\newblock {\em {Statistical Mechanics of Learning}}.
\newblock Cambridge University Press, 2001.

\bibitem{Zdeborova2016}
L.~Zdeborov{\'{a}} and F.~Krzakala.
\newblock {Statistical physics of inference: thresholds and algorithms}.
\newblock {\em Adv. Phys.}, 65(5):453--552, 2016.

\bibitem{Advani2016}
M.~S. Advani and S.~Ganguli.
\newblock {Statistical mechanics of optimal convex inference in high
  dimensions}.
\newblock {\em Physical Review X}, 6(3):1--16, 2016.

\bibitem{Chaudhari2017}
P.~Chaudhari, A.~Choromanska, S.~Soatto, Y.~LeCun, C.~Baldassi, C.~Borgs,
  J.~Chayes, L.~Sagun, and R.~Zecchina.
\newblock {Entropy-SGD: Biasing Gradient Descent Into Wide Valleys}.
\newblock In {\em ICLR}, 2017.

\bibitem{Advani2017}
M.~Advani and A.~M. Saxe.
\newblock {High-dimensional dynamics of generalization error in neural
  networks}.
\newblock {\em arXiv:1710.03667}, 2017.

\bibitem{Aubin2018}
B.~Aubin, A.~Maillard, J.~Barbier, F.~Krzakala, N.~Macris, and
  L.~Zdeborov{\'{a}}.
\newblock {The committee machine: Computational to statistical gaps in learning
  a two-layers neural network}.
\newblock In {\em Advances in Neural Information Processing Systems 31}, pages
  3227--3238, 2018.

\bibitem{Baity-Jesi2018}
M.~Baity-Jesi, L.~Sagun, M.~Geiger, S.~Spigler, G.B. Arous, C.~Cammarota,
  Y.~LeCun, M.~Wyart, and G.~Biroli.
\newblock {Comparing Dynamics: Deep Neural Networks versus Glassy Systems}.
\newblock In {\em Proceedings of the 35th International Conference on Machine
  Learning}, 2018.

\bibitem{Schwarze1993a}
H.~Schwarze.
\newblock {Learning a rule in a multilayer neural network}.
\newblock {\em Journal of Physics A: Mathematical and General},
  26(21):5781--5794, 1993.

\bibitem{Cybenko1989}
G.~Cybenko.
\newblock {Approximation by superpositions of a sigmoidal function}.
\newblock {\em Math. Control. Signals Syst.}, 2(4):303--314, 1989.

\bibitem{Hornik1989}
K.~Hornik, M.~Stinchcombe, and H.~White.
\newblock {Multilayer feedforward networks are universal approximators}.
\newblock {\em Neural Networks}, 2(5):359--366, 1989.

\bibitem{Mei2018}
S.~Mei, A.~Montanari, and P.-M. Nguyen.
\newblock {A mean field view of the landscape of two-layer neural networks}.
\newblock {\em Proceedings of the National Academy of Sciences},
  115(33):E7665--E7671, 2018.

\bibitem{Rotskoff2018}
G.~M. Rotskoff and E.~Vanden-Eijnden.
\newblock {Parameters as interacting particles: long time convergence and
  asymptotic error scaling of neural networks}.
\newblock In {\em Advances in neural information processing systems 31}, pages
  7146--7155, 2018.

\bibitem{Chizat2018}
L.~Chizat and F.~Bach.
\newblock On the global convergence of gradient descent for over-parameterized
  models using optimal transport.
\newblock In {\em Advances in Neural Information Processing Systems 31}, pages
  3040--3050, 2018.

\bibitem{Li2018a}
Y.~Li and Y.~Liang.
\newblock {Learning Overparameterized Neural Networks via Stochastic Gradient
  Descent on Structured Data}.
\newblock In {\em Advances in Neural Information Processing Systems 31}, 2018.

\bibitem{Kinzel1990}
W.~Kinzel and P.~Ruj{\'{a}}n.
\newblock {Improving a Network Generalization Ability by Selecting Examples}.
\newblock {\em EPL (Europhysics Letters)}, 13(5):473--477, 1990.

\bibitem{Mace1998}
C.W.H. Mace and A.C.C. Coolen.
\newblock {Statistical mechanical analysis of the dynamics of learning in
  perceptrons}.
\newblock {\em Statistics and Computing}, 8(1):55--88, 1998.

\bibitem{Saad1997b}
D.~Saad and S.A. Solla.
\newblock {Learning with Noise and Regularizers Multilayer Neural Networks}.
\newblock In {\em Advances in Neural Information Processing Systems 9}, pages
  260--266, 1997.

\bibitem{Oja1985}
E.~Oja and J.~Karhunen.
\newblock {On stochastic approximation of the eigenvectors and eigenvalues of
  the expectation of a random matrix}.
\newblock {\em Journal of Mathematical Analysis and Applications},
  106(1):69--84, 1985.

\bibitem{Wang2017}
C.~Wang, J.~Mattingly, and Yue~M. Lu.
\newblock {Scaling Limit: Exact and Tractable Analysis of Online Learning
  Algorithms with Applications to Regularized Regression and PCA}.
\newblock {\em arXiv:1712.04332}, 2017.

\bibitem{Wang2018}
C.~Wang, H.~Hu, and Y.~M. Lu.
\newblock {A Solvable High-Dimensional Model of GAN}.
\newblock {\em arXiv:1805.08349}, 2018.

\bibitem{Brutzkus2018}
A.~Brutzkus, A.~Globerson, E.~Malach, and S.~Shalev-Shwartz.
\newblock {SGD} learns over-parameterized networks that provably generalize on
  linearly separable data.
\newblock In {\em International Conference on Learning Representations}, 2018.

\bibitem{Soltanolkotabi2018}
M.~Soltanolkotabi, A.~Javanmard, and J.D. Lee.
\newblock {Theoretical insights into the optimization landscape of
  over-parameterized shallow neural networks}.
\newblock {\em IEEE Transactions on Information Theory}, 65(2):742--769, 2018.

\bibitem{Krogh1992a}
A.~Krogh and J.~A. Hertz.
\newblock {Generalization in a linear perceptron in the presence of noise}.
\newblock {\em Journal of Physics A: Mathematical and General},
  25(5):1135--1147, 1992.

\bibitem{Saxe2014}
A.M. Saxe, J.L. McClelland, and S.~Ganguli.
\newblock {Exact solutions to the nonlinear dynamics of learning in deep linear
  neural networks}.
\newblock In {\em ICLR}, 2014.

\bibitem{Lampinen2018}
A.K. Lampinen and S.~Ganguli.
\newblock An analytic theory of generalization dynamics and transfer learning
  in deep linear networks.
\newblock In {\em International Conference on Learning Representations}, 2019.

\end{thebibliography}

\end{document}